\RequirePackage[T1]{fontenc}
\documentclass[journal]{IEEEtran}

\usepackage{makecell}
\usepackage{graphicx}

\usepackage{amsmath,amsfonts,adjustbox}

\usepackage{algorithm,algpseudocode}

\usepackage{stfloats}

\usepackage{gensymb}

\usepackage[hidelinks]{hyperref}

\begin{document}

\title{KAUSTics-Net: Diffusion-Based Through-Water UAV Restoration with a Semi-Synthetic Dataset}

\author{Dario~Scilla$^{1}$,
        Omar~A.~Lopez$^{1}$,
        Kasper~Johansen$^{1}$,
        Brian~O.~Nieuwenhuis$^{2}$,
        and Matthew~F.~McCabe$^{1}$%
\thanks{$^{1}$Biological and Environmental Science and Engineering Division, 
Climate and Livability Initiative, King Abdullah University of Science and Technology, 
Thuwal 23955-6900, Saudi Arabia (e-mail: dario.scilla@kaust.edu.sa; 
omar.lopezcamargo@kaust.edu.sa; kasper.johansen@kaust.edu.sa; matthew.mccabe@kaust.edu.sa).}%
\thanks{$^{2}$Red Sea Research Center, Biological and Environmental Science and Engineering Division, 
King Abdullah University of Science and Technology, Thuwal 23955-6900, Saudi Arabia 
(e-mail: brian.nieuwenhuis@kaust.edu.sa).}%
}%

\maketitle

\begin{abstract}

Accurate mapping of shallow-water environments using unmanned aerial vehicle (UAV) imagery is challenged by water-induced distortions, including refraction, caustics, and wavelength-dependent attenuation. The scarcity of paired distorted and reference imagery further limits the development of supervised restoration methods. To address this challenge, we introduce KAUSTics, a semi-synthetic dataset combining 1,280 reference patches derived from 32 high-resolution UAV images with 12,800 degraded observations generated through Blender-based water-surface rendering and empirical color and blur augmentations. We further propose KAUSTics-Net, a pixel-space conditional diffusion framework integrating a multi-scale conditional encoder, feature-wise linear modulation (FiLM), cross-attention, and a composite reconstruction objective to restore shallow-water imagery. We evaluated the framework against eight image-restoration methods, complemented by architectural ablations and cross-dataset experiments. On KAUSTics, KAUSTics-Net achieved a mean peak signal-to-noise ratio (PSNR) of $31.715 \pm 0.439$~dB, structural similarity index (SSIM) of $0.849 \pm 0.002$, and learned perceptual image patch similarity (LPIPS) of $0.145 \pm 0.007$ across three training seeds. On the independent Sea-Undistort dataset, sampler-aware fine-tuning improved restoration fidelity, reaching $19.469 \pm 0.332$~dB PSNR and $0.554 \pm 0.005$ SSIM under combined water-surface distortions, although perceptual trade-offs remained. Additional experiments on real UAV imagery examined feature preservation and segmentation behavior. These findings demonstrate the potential of semi-synthetic training and conditional diffusion for shallow-water image restoration while highlighting the importance of evaluating both reconstruction fidelity and downstream remote-sensing utility.
\end{abstract}


\section{Introduction}

\IEEEPARstart{C}oastal regions support critical ecosystems, including coral reefs, seagrass meadows, mangroves, and tidal flats, while also hosting substantial populations and infrastructure. Their monitoring is essential for ecosystem assessment, coastal management, restoration, and climate adaptation \cite{jmse12030386,MOTTAZANIN2024120667,deguenon_sea-level_2024,CHOWDHURY202575}. Although in situ surveys provide detailed observations, they are spatially limited, logistically demanding, and often unable to capture rapid changes in dynamic coastal environments \cite{holzner_analyzing_2025,rs9070653,TSAI2024107136}. Satellite and unmanned aerial vehicle (UAV) remote sensing provide broader and more frequent coverage, enabling shallow-water mapping at scales impractical for field surveys alone \cite{app15094771,YANG2022155939,KUJAWA2025104604}.

Extracting reliable information from optical imagery over shallow coastal waters nevertheless remains challenging. Surface waves focus and defocus sunlight, producing dynamic, high-frequency caustic patterns unrelated to actual seabed variability \cite{8391752,scilla2025computer}. Specular reflection at the air--water interface generates sun glint that can saturate parts of the image and obscure the underlying scene \cite{QIN2024298,rs11202422}. Refraction displaces apparent object locations and distorts submerged shapes, while wavelength-dependent absorption and scattering alter the signal returning from the seafloor. Together, these effects degrade contrast and color, distort geometry, and complicate applications such as benthic habitat mapping \cite{rs13214452}, bathymetry \cite{agrafiotis2020correcting}, and photogrammetry \cite{Partama_photogrammetry}.

Several approaches have been proposed to mitigate water-induced distortions. Physics-based methods invert simplified radiative-transfer models but often require estimates of water depth, optical properties, and atmospheric conditions \cite{maritorena_diffuse_1994,dierssen_ocean_2003,rs12111861}. Image-processing approaches apply deglinting, dehazing, contrast enhancement, or multi-view corrections \cite{Hedley01052005,agrafiotis2020correcting,scilla2025computer}. More recently, deep learning has enabled data-driven mappings between distorted and reference imagery. GAN-based methods have been applied to underwater enhancement and coastal water removal \cite{9570386,QIN2024298,10720144}, while diffusion-based image-to-image models have demonstrated capabilities for removing complex degradations while preserving scene structure \cite{10.1145/3588432.3591513,Xia_2023_ICCV}.

However, learning-based restoration of shallow-water UAV imagery is constrained by the scarcity of large, high-resolution paired datasets containing observations of the same scene with and without water-surface distortions \cite{10720144,kromer2025sea}. Repeated acquisitions under different sea states and illumination rarely provide aligned pairs and make it difficult to separate water-surface effects from genuine scene changes \cite{KUJAWA2025104604}. Purely synthetic data, in contrast, may not capture the spatial complexity and variability of real coastal environments. This motivates semi-synthetic approaches that combine real UAV imagery with physically based simulations of water-induced distortions.

To address this gap, we introduce KAUSTics, a semi-synthetic dataset for high-resolution UAV water-distortion removal, together with KAUSTics-Net, a conditional diffusion-based restoration framework. We evaluated the proposed approach on KAUSTics and the independent Sea-Undistort benchmark \cite{kromer2025sea}, complemented by an exploratory assessment on real UAV imagery.

In summary, this work makes the following contributions:
\begin{itemize}
    \item We introduce KAUSTics, a semi-synthetic dataset comprising 1,280 high-resolution UAV reference patches and 12,800 degraded observations generated through physically based water-surface rendering and empirical color and blur augmentations.

    \item We propose KAUSTics-Net, a pixel-space conditional diffusion framework combining multi-scale conditioning, FiLM-modulated residual blocks, cross-attention, and a composite restoration objective for through-water UAV imagery.

    \item We evaluate KAUSTics-Net against eight restoration methods using source-image-disjoint partitions, architectural ablations, cross-dataset experiments, and real-UAV assessments of downstream image-processing utility.
\end{itemize}

\section{Related Work}
\label{sec:related_work}

Image restoration has been extensively studied across computer vision and remote sensing, including dehazing, deblurring, deraining, and underwater image enhancement. However, methods explicitly targeting water-surface distortions such as caustics, sun glint, and refraction in high-resolution UAV imagery remain comparatively limited. In practice, many approaches applied to this problem were originally developed for generic image restoration or underwater cameras rather than through-water aerial observations \cite{fabbri2018enhancing,10720144,guan2023diffwater}. Existing approaches can be broadly grouped into statistical and physics-based corrections, generative adversarial networks (GANs), and diffusion-based restoration models.

\subsection{Statistical and Physics-Based Approaches}

In UAV-based shallow-water imaging, refraction at the air--water interface displaces apparent object locations and distorts submerged shapes, while wave-induced caustics and sun glint introduce strong spatially varying radiometric artifacts \cite{Chong03082022,scilla2025computer}. Early correction methods primarily focused on geometric effects by applying empirical refraction factors derived from Snell's law. Partama \emph{et al.} proposed a correction of apparent depths from UAV structure-from-motion reconstructions, while related studies incorporated refraction correction into shallow-water bathymetric inversion schemes \cite{partama2017simple,lambert2023refraction}. Reviews of UAV-based shallow-water bathymetry have highlighted the effectiveness of these approaches, but also their dependence on acquisition geometry, water depth, seabed properties, and assumptions regarding the water surface \cite{luis2015synoptic,specht2022analysis}.

A complementary family of methods exploits temporal redundancy in high-frame-rate imagery. Fluid lensing uses repeated observations of the same scene to model the dynamic water surface as a time-varying optical lens and reconstruct refraction-corrected, caustic-reduced imagery \cite{chirayath2016drones,chirayath2019fluid}. More recently, Scilla \emph{et al.} showed that frame-wise temporal aggregation combined with color-transfer correction can reduce caustics, refraction-induced shape variations, and sun glint in UAV video while improving the consistency of submerged features \cite{scilla2025computer}.

Although effective, these approaches generally rely on explicit physical assumptions, hand-crafted corrections, or repeated observations of the same scene. They therefore do not directly learn a single-image mapping from a distorted observation to a clean seabed representation, motivating the use of data-driven generative models.

\subsection{Generative Adversarial Networks}

Generative adversarial networks learn image distributions through competition between a generator and discriminator \cite{goodfellow2014generative}, and have been widely adopted for image enhancement and restoration. In image-to-image translation, CycleGAN introduced cycle consistency for learning mappings between unpaired image domains \cite{zhu2017unpaired}, a particularly useful property when paired distorted and clean observations are difficult to acquire.

Underwater enhancement models such as UGAN, FUnIE-GAN, MLFcGAN, and UWGAN use adversarial learning to improve color balance, contrast, local structure, and, in some cases, downstream robotic perception \cite{fabbri2018enhancing,islam2020fast,liu2019mlfcgan,wang2019uwgan}. These approaches have demonstrated substantial improvements for imagery acquired by underwater cameras. However, their target domain differs from through-water UAV imagery, where wave-induced refraction, dynamic caustics, and sun glint introduce stronger spatially varying geometric and radiometric distortions.

Only a limited number of GAN-based methods have directly addressed aerial water removal. DeWaterGAN is a physics-guided unsupervised framework for UAV coastal imagery that combines a CycleGAN backbone with a light-transport-inspired attention mechanism \cite{10720144}. The model can reduce water interference while preserving color and texture, but it primarily targets appearance restoration and does not rely on pixel-aligned distorted/undistorted training pairs.

Overall, GAN-based methods demonstrate the potential of learned restoration, but their application to through-water UAV imagery remains challenging, as these images belong to substantially different domains, containing stronger refraction induced distortions, and complex caustic patterns that are not fully captured in classical underwater datasets.

\subsection{Diffusion and Advanced Restoration Models}

Denoising diffusion probabilistic models progressively corrupt training data with Gaussian noise and learn the reverse denoising process \cite{ho2020denoising}. Latent diffusion models, including Stable Diffusion, further reduce computational cost by performing this process in a compressed representation space \cite{rombach2022high}. Conditional diffusion models extend this framework to image-to-image restoration by incorporating degraded observations into the denoising process. For instance, Palette introduced a general conditional diffusion framework for image-to-image translation \cite{saharia2022palette}.

Several diffusion-based approaches have been developed for underwater image enhancement. UW-DDPM introduced a dual-U-Net formulation with distribution-fusion sampling \cite{lu2023ddpmuiew}, while DiffWater incorporated degraded-image and color-compensated guidance to improve underwater image restoration \cite{guan2023diffwater}. Related studies have investigated accelerated sampling strategies and alternative conditioning mechanisms \cite{lu2023speed,shi2024cpdm}. Beyond underwater imagery, NDR-Restore learns degradation representations to address multiple corruption types \cite{yao2024neural}, while ResShift formulates diffusion in the residual space between degraded and reference images to reduce the number of sampling steps \cite{yue2023resshift}.

Alternative deep learning architectures have also been developed for underwater image enhancement. SS-UIE combines spatial selective scanning and spectral attention to address nonuniform image degradation \cite{peng_ssuie_2025}, while U-Shape Transformer integrates multi-scale channel-wise feature fusion and spatial global modeling \cite{peng2023u}. These methods provide complementary restoration strategies beyond conventional convolutional and diffusion-based frameworks.

More recently, the Sea-Undistort dataset introduced a paired synthetic benchmark for through-water airborne imagery and evaluated restoration models under sun glint, scattering, and wave-induced distortions \cite{kromer2025sea}. These results demonstrate the potential of learning-based restoration for aerial through-water observations. However, most existing restoration methods were developed for underwater-camera imagery or general image degradation rather than through-water UAV observations. Moreover, the availability of large paired datasets preserving the texture and spatial complexity of real coastal UAV imagery remains limited.

The present work addresses this gap by introducing KAUSTics, a semi-synthetic dataset combining real high-resolution UAV imagery with rendered water-surface distortions and empirical appearance augmentations, together with KAUSTics-Net, a conditional diffusion framework tailored to this setting. This combination enables supervised restoration toward calm-water reference imagery and controlled evaluation under multiple caustic patterns, blur levels, and color transformations.

\section{Methodology}
\subsection{KAUSTics Dataset}
\label{sec:dataset}

\begin{figure}[!h]
\centering
\includegraphics[width=0.82\linewidth]{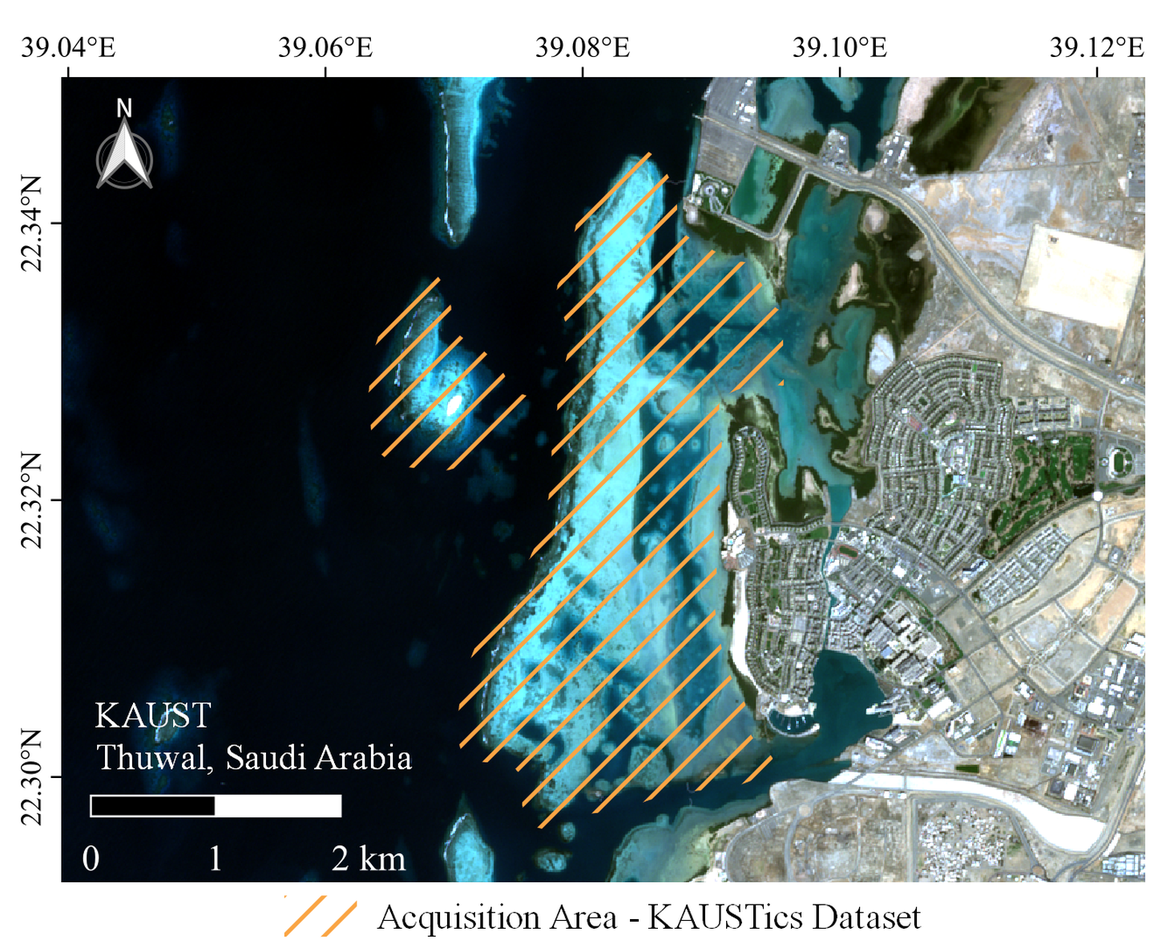}
\caption{Area of interest for the KAUSTics dataset. The background shows a Sentinel-2 RGB composite of the coastal waters near Thuwal, Saudi Arabia, in the central Red Sea. The orange lines indicate the approximate area of interest of the UAV surveys used to acquire the calm-water reference imagery over shallow reef flats and sandy substrates.}
\label{fig:aoi_kaustics}
\end{figure}

The KAUSTics dataset builds on UAV surveys conducted in June 2024 between 8 and 9~AM at the King Abdullah University of Science and Technology (KAUST), approximately 80~km north of Jeddah on the Red Sea coast of Saudi Arabia (22.34$^\circ$N, 39.09$^\circ$E). The surveys were originally collected for coastal monitoring applications, including habitat mapping. For this study, we selected 32 images acquired over a shallow-water lagoon (depth $<3$~m; Fig.~\ref{fig:aoi_kaustics}) under clear-sky and low-wind conditions to minimize visible water-surface distortions.

The images were acquired using a DJI Matrice 300 RTK flying at 35~m altitude and equipped with a Zenmuse P1 45-MP full-frame camera with a 35~mm lens. Each frame had a resolution of $8192 \times 5460$~pixels and a ground sampling distance of approximately $0.4~cm/pixel$. The surveyed area included sandy, rocky, seagrass, and macroalgal substrates. The selected images represent different seabed textures and visual characteristics observed in the lagoon and serve as calm-water reference targets. Their geometric and radiometric accuracy has not been independently established.

\begin{figure}[!h]
\centering
\includegraphics[width=0.82\linewidth]{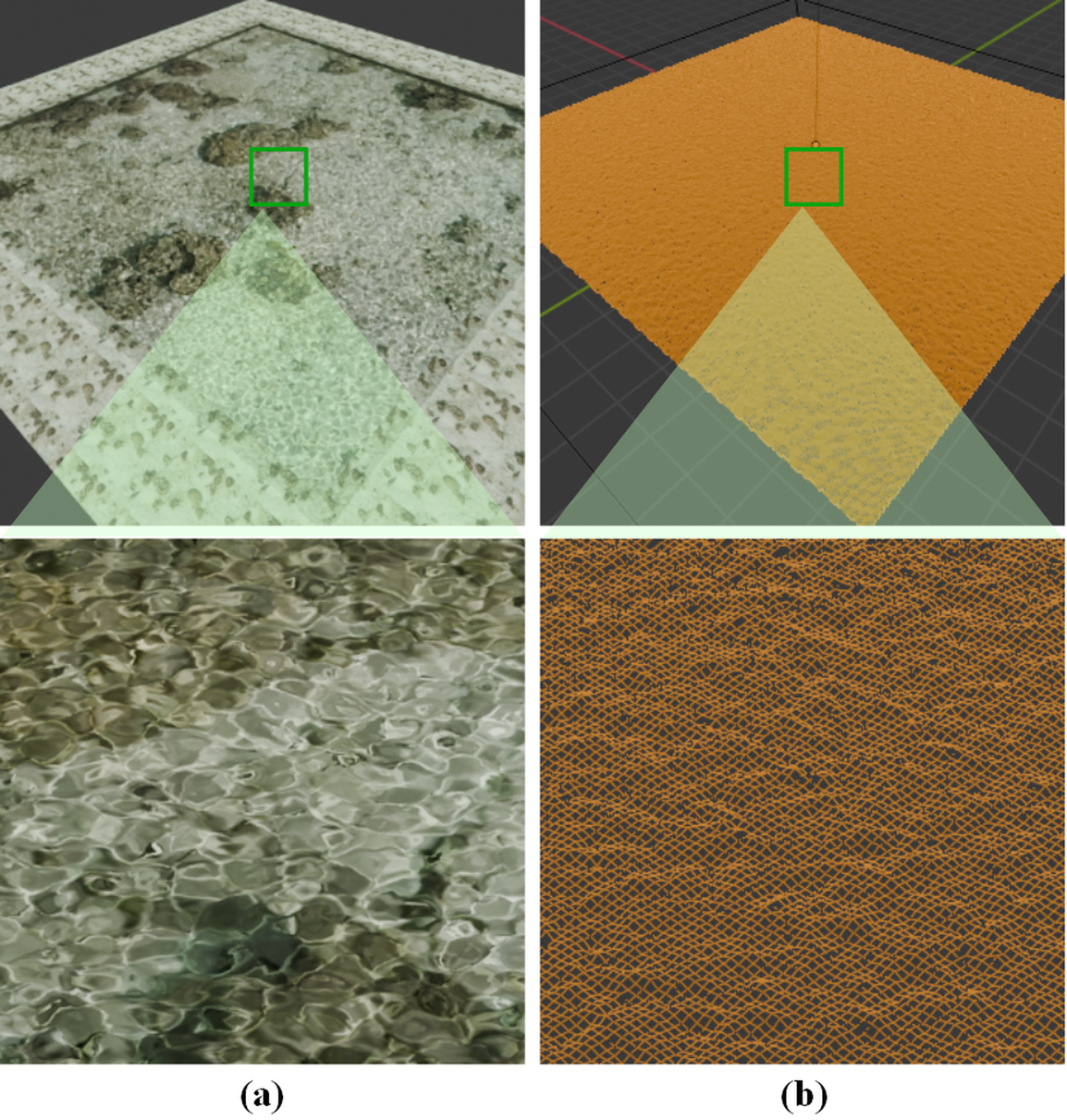}
\caption{Rendering setup used to generate the synthetic component of KAUSTics. (a) Blender scene showing the UAV-derived seafloor texture mapped onto the bottom plane and viewed by a nadir-looking camera. (b) Mesh of the simulated water surface generated using the Ocean Modifier, with a central zoom showing its spatial sampling density.}
\label{fig:blender}
\end{figure}

Synthetic surface distortions were generated using Blender \cite{blender} by mapping the UAV-derived reference imagery onto the bottom plane of a virtual shallow-water scene and rendering it through a simulated water surface. The rendering process introduced wave-induced illumination focusing (caustics) and refractive displacement. Sun glint was not independently parameterized or controlled in the KAUSTics rendering process. The virtual camera was configured to provide a reproducible nadir-viewing geometry at an altitude of 10~m, with a sensor size of $13.2 \times 8.8$~mm, a focal length of 6.201~mm, and a rendering resolution of $1000 \times 1000$~pixels. The rendering setup is illustrated in Fig.~\ref{fig:blender}.

\begin{figure*}[!t]
\centering
\includegraphics[width=0.94\linewidth]{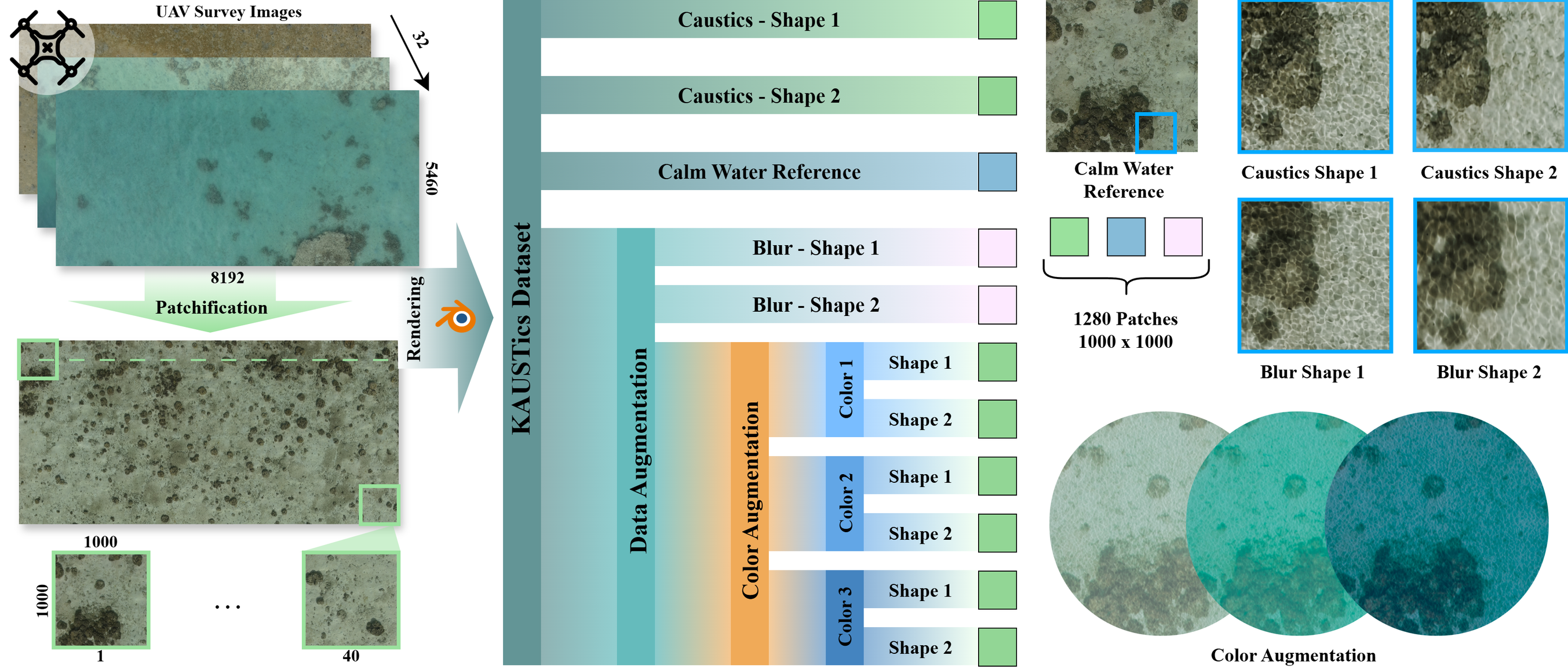}
\caption{Overview of the KAUSTics semi-synthetic dataset. UAV imagery is divided into $1000 \times 1000$~pixel reference patches and rendered in Blender to generate two caustic patterns. Gaussian blur and three empirical color-transfer configurations (Color~1--3) expand the dataset to ten degraded observations per reference patch.}
\label{fig:dataset}
\end{figure*}

Water-surface geometry was simulated using Blender's Ocean Modifier with the \textit{Shallow Water} spectrum, a depth parameter of 0.4~m, Spatial Size of 49~m, wave-amplitude Scale of 0.025, and wind velocity of 5~m~s\textsuperscript{-1}. The geometry Size parameter was set to 0.190, with a water-object scale of approximately 1.665. Two distinct caustic patterns were generated by varying the surface alignment parameters: Shape~1 used an alignment of 0, producing predominantly cell-like caustics, whereas Shape~2 used an alignment of 0.1, a direction of 90\degree, and damping of 0.5, producing more parallel structures (Fig.~\ref{fig:dataset}). Surface realizations were varied through the Ocean Modifier's random seed. All scenes were rendered using Blender Cycles with OptiX acceleration on an NVIDIA A100 GPU (80~GB). These parameters define the synthetic scene configuration and do not represent measured hydrodynamic or optical properties of the lagoon.

The resulting dataset follows the hierarchical organization illustrated in Fig.~\ref{fig:dataset}. Starting from the rendered caustic patterns, Gaussian-blurred variants were generated to approximate reduced visibility and optical defocus. Three additional color transformations (Color~1--3) were applied using statistical color transfer in CIELAB space \cite{reinhard2002color}, matching channel means and standard deviations to manually selected target appearances. These transformations introduce variability in spectral appearance and serve as empirical proxies for absorption- and turbidity-like effects. They do not constitute calibrated simulations of water-column optical properties or measured depth-dependent attenuation.

Each of the 32 UAV images contributed 40 reference patches of $1000 \times 1000$~pixels, resulting in 1,280 reference patches. Each reference was associated with ten degraded observations: two caustic patterns, two Gaussian-blurred variants, and six color-transfer variants corresponding to three color configurations applied to both caustic patterns. The complete dataset therefore comprises 12,800 distorted--reference pairs. These augmentations increase the diversity of simulated degradation appearances without increasing the number of independent source images. The source-image-disjoint training, validation, and test partitioning protocol is described in Section~\ref{sec:experimental_setup}.

Although each degraded observation is associated with its corresponding reference patch, wave-induced refraction can introduce local geometric displacement, preventing exact feature-wise pixel correspondence. The calm-water references were acquired through water and may retain residual refraction and radiometric attenuation. No independent water-free seabed images, surveyed geometric references, or calibrated reflectance measurements are available. Consequently, KAUSTics supports supervised restoration toward a low-distortion, calm-water reference appearance. Additional dataset visualizations and rendering specifications are provided in the supplementary material.

\subsection{Sea-Undistort Dataset}
\label{sec:su_dataset}

To assess the generalization of KAUSTics-Net beyond the KAUSTics dataset, we also evaluated it on Sea-Undistort \cite{kromer2025sea}, a synthetic through-water benchmark developed in Blender for airborne bathymetric mapping. The dataset comprises 1,200 scenes, originally rendered as $512 \times 512$~pixel RGB images. The simulated seabed includes randomized combinations of sandy flats, gravel, rocky substrates, and seagrass-like vegetation, providing a range of textures and reflectance characteristics.

\begin{figure}[!h]
\centering
\includegraphics[width=0.85\linewidth]{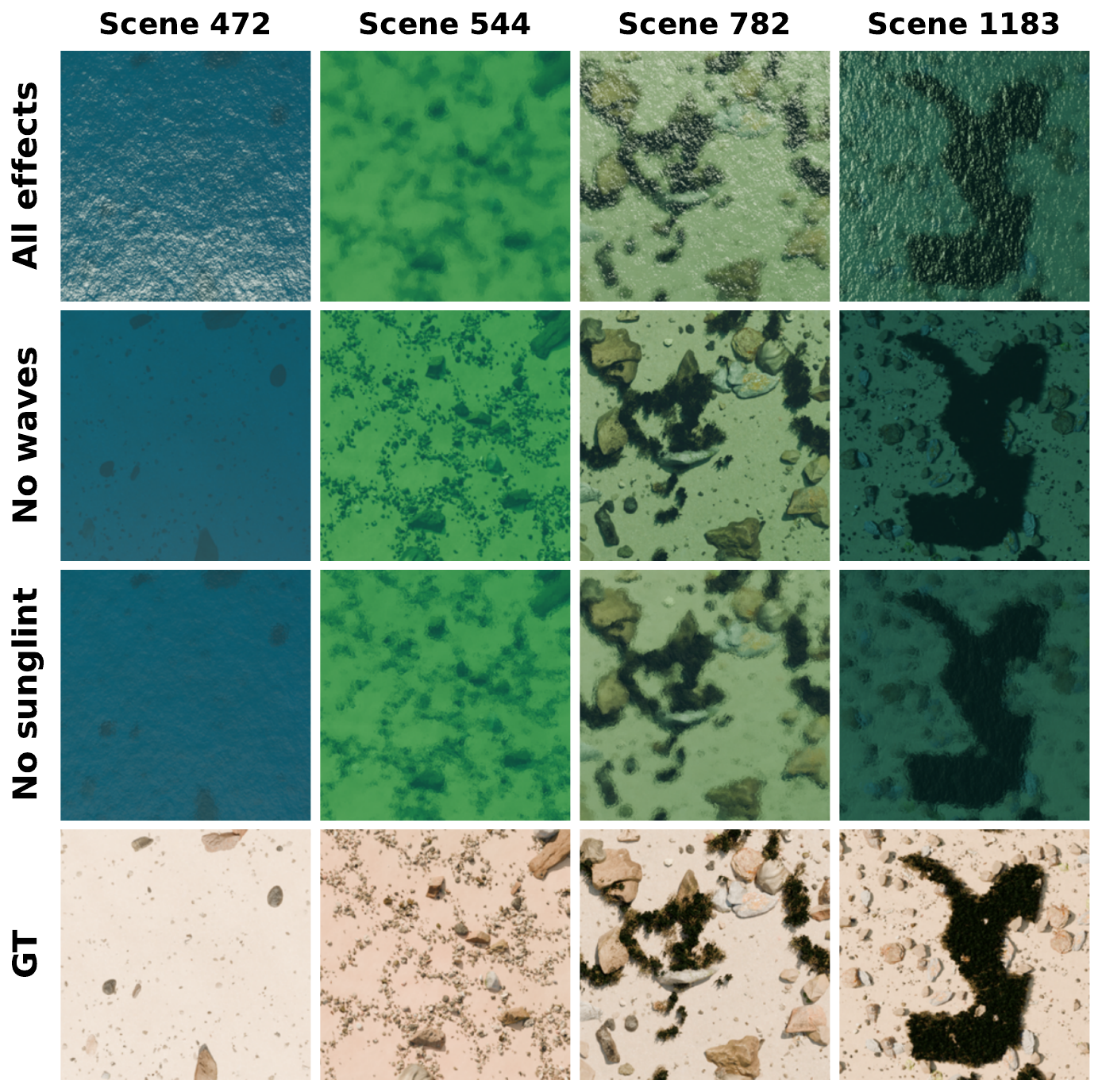}
\caption{Examples from the Sea-Undistort dataset illustrating four scenes under three distortion configurations: all effects, no waves, and no sun glint. The bottom row shows the corresponding original water-free reference images, which are identical across the three configurations.}
\label{fig:dataset_SU}
\end{figure}

For each scene, Sea-Undistort provides an original water-free reference image and multiple distorted observations generated under different water-surface and water-column conditions, including refraction, sun glint, light absorption, and turbidity-related effects. In this study, we considered three distortion configurations: \textit{all effects}, \textit{no waves}, and \textit{no sunglint}. The all-effects configuration combines surface-wave and water-column distortions, while the other two configurations exclude the corresponding surface effect. All three share the same water-free reference image, enabling consistent evaluation across distortion conditions. Some observations exhibit strong glint and wave-induced distortions that partially or completely obscure the seabed, increasing the difficulty of restoration. Figure~\ref{fig:dataset_SU} illustrates representative scenes under the three configurations and their corresponding water-free references.

Table~\ref{tab:dataset_distortions} summarizes the distortions represented in both datasets. KAUSTics includes rendered caustics and refraction, while color-transfer and Gaussian-blur augmentations approximate absorption- and turbidity-related appearance changes. Sea-Undistort provides rendered refraction, sun glint, and water-column effects. Caustics were not independently controlled in the Sea-Undistort configurations considered here. The two datasets therefore provide complementary conditions for assessing restoration performance.

\begin{table}[!h]
\centering
\caption{Distortions represented in KAUSTics and Sea-Undistort, distinguishing physically rendered effects from empirical data augmentations.}
\label{tab:dataset_distortions}
\footnotesize
\setlength{\tabcolsep}{4pt}
\begin{tabular}{lcc}
\hline
\textbf{Distortion} & \textbf{KAUSTics} & \makecell{\textbf{Sea-Undistort}} \\
\hline
Caustics         & Rendered  & --- \\
Refraction       & Rendered  & Rendered \\
Sun glint        & ---       & Rendered \\
Light absorption & Augmented & Rendered \\
Turbidity        & Augmented & Rendered \\
\hline
\end{tabular}
\end{table}

For the revised experiments, model adaptation used the all-effects observations, while testing also included the no-waves and no-sunglint configurations. The same scene identifiers and original water-free reference images were used across all three test conditions. The complete evaluation and adaptation protocols are described in the experimental setup section~\ref{sec:experimental_setup}.

\subsection{KAUSTics-Net Architecture}
\label{sec:method_arch}

\begin{figure*}
\centering
\includegraphics[width=0.92\linewidth]{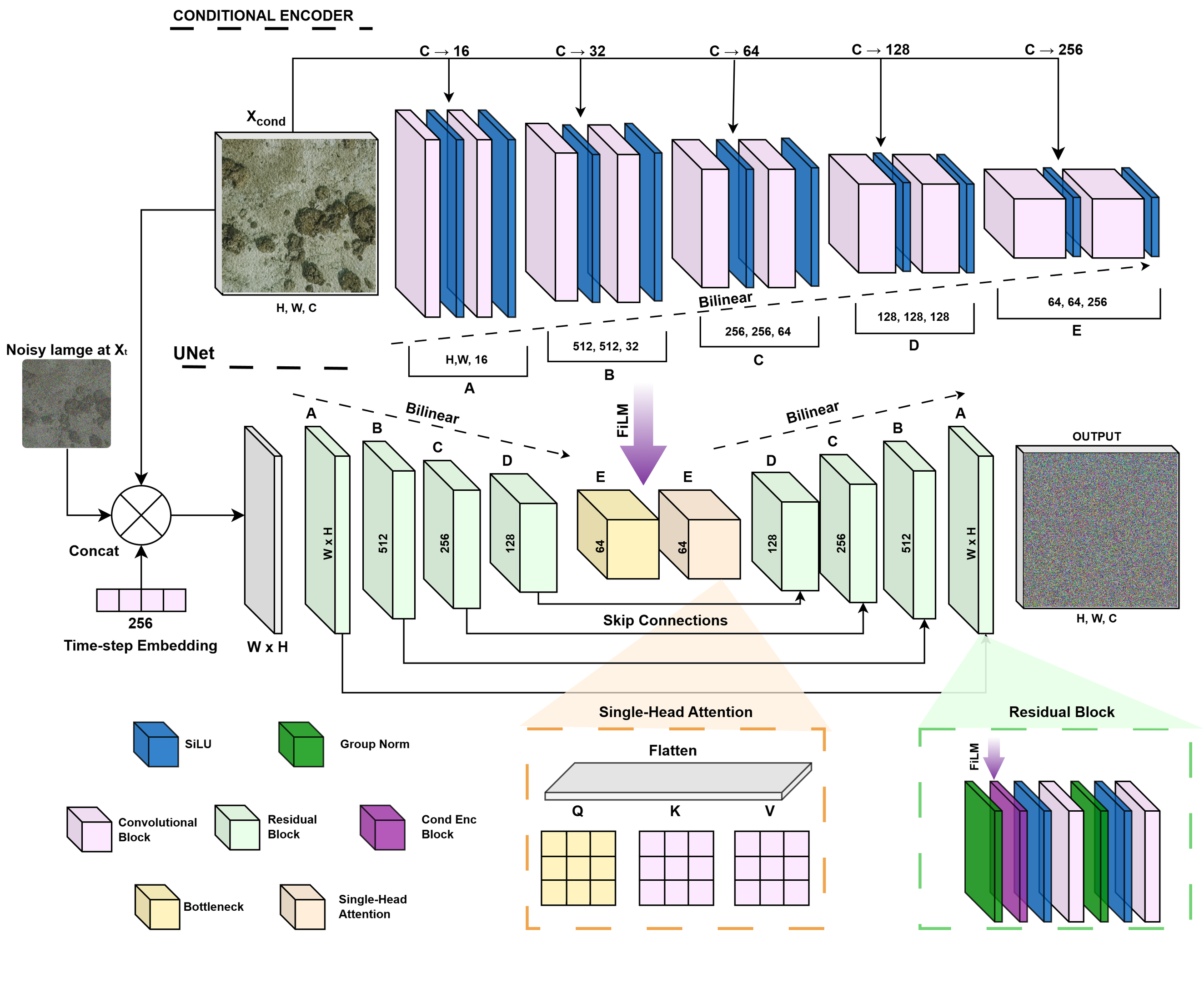}
\caption{Architecture of KAUSTics-Net. The distorted observation $\mathbf{x}_{\mathrm{cond}}$ is processed by a multi-scale conditional encoder and concatenated with the noisy target $\mathbf{x}_t$ at the U-Net input. Time- and image-conditioned FiLM residual blocks integrate conditioning information throughout the network, while single-head cross-attention fuses U-Net and conditional features at the bottleneck. The decoder predicts diffusion velocity $\mathbf{v}_{\theta}(\mathbf{x}_t,\mathbf{x}_{\mathrm{cond}},t)$ for reference-image reconstruction.}
\label{fig:diff_model}
\end{figure*}

KAUSTics-Net is a pixel-space conditional diffusion framework designed to restore shallow-water UAV imagery affected by spatially varying water-induced distortions. The architecture consists of a U-Net denoiser, a multi-scale conditional encoder, feature-wise linear modulation (FiLM) residual blocks, and a cross-attention bottleneck (Fig.~\ref{fig:diff_model}). The model operates directly on RGB images without a latent autoencoder or pretrained diffusion components and predicts diffusion velocity conditioned on the distorted observation.

Let $\mathbf{x}_0 \in \mathbb{R}^{H \times W \times 3}$ denote the reference image and $\mathbf{x}_{\mathrm{cond}} \in \mathbb{R}^{H \times W \times 3}$ its corresponding distorted observation. During training, a noisy sample $\mathbf{x}_t$ is generated from $\mathbf{x}_0$ at diffusion timestep $t$, as described in Section~\ref{sec:method_forward}. The U-Net receives the channel-wise concatenation of $\mathbf{x}_t$ and $\mathbf{x}_{\mathrm{cond}}$, while a separate five-stage conditional encoder extracts multi-scale features from the distorted observation. The encoder uses $3 \times 3$ convolutions, group normalization, SiLU activations, and bilinear downsampling, with channel dimensions of 16, 32, 64, 128, and 256.

The extracted features condition the U-Net residual blocks through FiLM \cite{perez2018film}. A 256-dimensional sinusoidal timestep embedding and globally pooled conditional features are projected into channel-wise scale and shift parameters, allowing the residual blocks to incorporate both the diffusion timestep and image-dependent information. At the $16 \times 16$ bottleneck, a single-head cross-attention mechanism with 256-dimensional features further integrates the conditioning information. U-Net features serve as queries, while projected conditional features provide keys and values. The decoder combines these representations with skip connections to predict the diffusion velocity $\mathbf{v}_{\theta}(\mathbf{x}_t,\mathbf{x}_{\mathrm{cond}},t)$.

This conditioning strategy is motivated by the combination of localized caustics and refractive distortions with broader radiometric changes in through-water imagery. Multi-scale features provide image-dependent information throughout denoising, while the bottleneck attention mechanism integrates spatial information from the distorted observation. The individual architectural components follow established diffusion and conditioning techniques; their contribution within KAUSTics-Net is examined through the ablation experiments.

\subsection{Diffusion Training and Image Reconstruction}
\label{sec:method_forward}

KAUSTics-Net follows a discrete-time diffusion process \cite{ho2020denoising} with a cosine noise schedule \cite{nichol2021improved}. Reference images are normalized to $[-1,1]$, and the forward process uses $T=1000$ diffusion states. At training, a timestep $t$ is uniformly sampled from $\{1,\ldots,T-1\}$, and Gaussian noise $\boldsymbol{\varepsilon}\sim\mathcal{N}(\mathbf{0},\mathbf{I})$ is added to the reference image $\mathbf{x}_0$:

\begin{equation}
\mathbf{x}_t
=
a_t\mathbf{x}_0+b_t\boldsymbol{\varepsilon},
\qquad
a_t=\sqrt{\bar{\alpha}_t},\quad
b_t=\sqrt{1-\bar{\alpha}_t},
\label{eq:q_sample}
\end{equation}
where $\bar{\alpha}_t=\prod_{s=1}^{t}(1-\beta_s)$ denotes the cumulative noise schedule.

The network predicts the velocity target, which combines the reference image and the sampled noise. The corresponding clean-image estimate is obtained as

\begin{equation}
\begin{aligned}
\mathbf{v}_t &= a_t\boldsymbol{\varepsilon}-b_t\mathbf{x}_0,\\
\hat{\mathbf{x}}_0 &= a_t\mathbf{x}_t-b_t\hat{\mathbf{v}}_{\theta},
\end{aligned}
\label{eq:velocity_pred}
\end{equation}
where $\hat{\mathbf{v}}_{\theta}=\mathbf{v}_{\theta}(\mathbf{x}_t,\mathbf{x}_{\mathrm{cond}},t)$ is the predicted velocity. This formulation enables direct reconstruction of the reference-image estimate during training and sampling.

At inference, the reverse diffusion process is initialized by adding Gaussian noise to the distorted observation $\mathbf{x}_{\mathrm{cond}}$, using a strength parameter of $s=0.95$, corresponding to diffusion state $t_0=949$. Restoration is performed using 25 deterministic DDIM sampling steps ($\eta=0$) \cite{song2020denoising}, progressing from the selected initial state toward the reconstructed image. The distorted observation is provided as conditioning at every step. Classifier-free guidance with a scale of 2 is applied using a conditioning dropout probability of 0.1 during training, resulting in 50 network evaluations per restored image.

The final reconstructed image is converted to the RGB intensity range for evaluation. The diffusion process and sampling configuration are shared across the KAUSTics experiments, while the Sea-Undistort adaptation protocol is described separately.

\subsection{Loss Functions and Optimization Strategy}
\label{sec:losses}

KAUSTics-Net was trained using a composite objective combining diffusion velocity prediction with spatial, structural, and chromatic reconstruction losses. The total objective is defined as
\begin{equation}
\begin{aligned}
\mathcal{L}_{\mathrm{total}} ={}&
0.2\,\mathcal{L}_{v}
+ 0.8\,\mathcal{L}_{\mathrm{RGB}}
+ 0.2\,\mathcal{L}_{\mathrm{LF}} \\
&+ 0.3\,\mathcal{L}_{\mathrm{UV}}
+ 0.2\,\mathcal{L}_{\mathrm{SSIM}}
+ 0.2\,\mathcal{L}_{\mathrm{Lab}},
\end{aligned}
\label{eq:ltotal}
\end{equation}
where the coefficients were selected empirically to balance reconstruction fidelity, structural consistency, and color preservation.

The velocity prediction loss $\mathcal{L}_{v}$ minimizes the mean squared error (MSE) between the predicted velocity $\hat{\mathbf{v}}_{\theta}$ and the target $\mathbf{v}_t$ defined in Eq.~\eqref{eq:velocity_pred}. The RGB reconstruction loss $\mathcal{L}_{\mathrm{RGB}}$ similarly applies MSE between the reconstructed image $\hat{\mathbf{x}}_0$ and the reference $\mathbf{x}_0$. To preserve large-scale appearance, the low-frequency loss $\mathcal{L}_{\mathrm{LF}}$ computes the $L_1$ difference between the reconstructed and reference images after Gaussian smoothing with a $21 \times 21$ kernel and standard deviation $\sigma=5$. The chromatic loss $\mathcal{L}_{\mathrm{UV}}$ applies $L_1$ differences to the U and V channels in YUV color space, encouraging consistent color reproduction.

Two additional terms constrain structural and perceptual color differences. The structural loss $\mathcal{L}_{\mathrm{SSIM}}$ is defined as one minus the structural similarity index (SSIM), computed over RGB channels using a uniform $7 \times 7$ window. The color-space loss $\mathcal{L}_{\mathrm{Lab}}$ measures the $L_1$ difference between Gaussian-smoothed images converted to D65/2-degree CIELAB space, with the L, a, and b channels normalized by 100, 128, and 128, respectively. Predicted RGB values are clipped to $[0,1]$ for the SSIM and CIELAB losses, while the RGB, low-frequency, and UV reconstruction losses use unclipped predictions. All loss terms are averaged over the image and batch. These objectives provide complementary reconstruction constraints without explicitly modeling water-column optical properties.

Optimization was performed using AdamW with a learning rate of $10^{-4}$, weight decay of $10^{-4}$, and momentum coefficients $(\beta_1,\beta_2)=(0.9,0.999)$. Training used a batch size of 12, FP32 precision, and gradient clipping at a global norm of 1.0. The model was trained for 1,000 epochs without learning-rate decay or exponential moving average (EMA) of network weights. Conditioning dropout was applied with probability $p_{\mathrm{cond}}=0.1$ to support classifier-free guidance during inference. Model checkpoints were selected using validation PSNR, as detailed in the experimental setup.

\subsection{Benchmark Models and Evaluation Metrics}
\label{sec:benchmark_metrics}

We benchmarked KAUSTics-Net against eight image-restoration methods spanning GAN-based architectures, underwater image enhancement, and diffusion-based restoration. The GAN-based baselines include CycleGAN \cite{zhu2017unpaired}, UGAN \cite{fabbri2018enhancing}, FUnIE-GAN \cite{islam2020fast}, and DeWaterGAN \cite{10720144}. The diffusion-based methods include DiffWater \cite{guan2023diffwater} and ResShift \cite{yue2023resshift}. We additionally evaluated SS-UIE \cite{peng_ssuie_2025} and U-Shape Transformer\cite{peng2023u}, which incorporate spatial--spectral modeling and transformer-based feature extraction, respectively. These methods provide complementary architectural comparisons for restoring imagery affected by spatially varying distortions. All eight baselines were trained and evaluated using the revised experimental protocol described in the following subsection.

Restoration fidelity was assessed using mean peak signal-to-noise ratio (mPSNR), mean structural similarity index (mSSIM), mean root mean squared error (mRMSE), and mean squared error (mMSE). We additionally included learned perceptual image patch similarity (LPIPS) \cite{zhang2018unreasonable} and CIEDE2000 ($\Delta E_{00}$) \cite{sharma2005ciede2000} to quantify perceptual feature differences and color discrepancies relative to the reference images. Higher values indicate better performance for mPSNR and mSSIM, while lower values are preferred for the remaining metrics. LPIPS was computed using an AlexNet backbone on 8-bit RGB outputs, and CIEDE2000 was evaluated in D65/2-degree CIELAB space.

No-reference image quality was evaluated using the Natural Image Quality Evaluator (NIQE) \cite{mittal2012making}, Blind/Referenceless Image Spatial Quality Evaluator (BRISQUE) \cite{mittal2011blind}, Perception-based Image Quality Evaluator (PIQE) \cite{venkatanath2015blind}, and Integrated Local NIQE (ILNIQE) \cite{zhang2015feature}. These metrics assess deviations from statistical properties associated with natural images, with lower values indicating better predicted quality. Three learning-based perceptual metrics were also included: Learned Image Quality Evaluator (LIQE) \cite{zhang2023blind}, CLIP-based Image Quality assessment (CLIPIQ) \cite{agnolucci2024qualityaware}, and Multi-scale Image Quality (MUSIQ) \cite{ke2021musiq}. For MUSIQ, we report the AVA-trained variant (MUSIQ\_AVA), with higher scores indicating better predicted perceptual quality.

Sharpness and Shannon entropy were further computed as complementary indicators of image detail and intensity variability. Sharpness was defined as the variance of the discrete Laplacian, while entropy was calculated from the 256-bin luminance histogram. These metrics characterize changes in image structure and tonal distribution but do not independently establish restoration accuracy, as noise and artifacts can also increase their values.

Finally, to investigate the potential utility of restoration for downstream UAV image analysis, we evaluated real-world examples using the Segment Anything Model (SAM) \cite{kirillov2023segany} and scale-invariant feature transform (SIFT) \cite{lowe2004distinctive}, following the computer-vision assessment approach previously applied to shallow-water UAV imagery \cite{scilla2025computer}. SAM was used to examine automatically generated image regions, while SIFT quantified local keypoint detection and descriptor correspondence. The real-UAV comparison used a temporal computer-vision reference from our previous work, which provides a low-distortion comparison image without constituting independently validated water-free ground truth. These analyses complement the image quality metrics by examining the preservation of local features and segmentation behavior.

\subsection{Experimental Setup}
\label{sec:experimental_setup}

The stored KAUSTics images were converted to RGB and normalized to $[-1,1]$ for training. All methods used the same $256 \times 256$ image resolution, obtained by bicubic downsampling of the original $1000 \times 1000$ patches without additional spatial cropping.

To prevent leakage between augmented observations of the same source image, the dataset was partitioned at the UAV-image level. Using a fixed random seed of 42, the 32 source images were divided into 26 training, three validation, and three test images. Every reference patch and its ten degraded variants inherited the partition of the corresponding source image. Table~\ref{tab:revision_partitions} summarizes the resulting dataset composition.

The partitioning protocol ensures independence of source-image identifiers across training, validation, and testing. However, geographic independence cannot be guaranteed because different UAV frames may cover overlapping seabed areas. For KAUSTics-Net, each training epoch contained 2,080 patch--shape combinations, with additional degradation variants sampled with probability 0.6. Model selection used a fixed, stratified subset of 120 validation observations evaluated through the conditional sampling procedure. The complete test set contained all 1,200 degraded observations from the three held-out UAV images.

\begin{table}[!h]
\centering
\caption{KAUSTics source-image splits. Each UAV image contributes 40 reference patches, with ten degraded observations per patch.}
\label{tab:revision_partitions}
\footnotesize
\begin{tabular}{lrrr}
\hline
\textbf{Partition} & \textbf{UAV images} & \textbf{Patches} & \textbf{Pairs} \\
\hline
Training   & 26 & 1,040 & 10,400 \\
Validation & 3  & 120   & 1,200  \\
Test       & 3  & 120   & 1,200  \\
\hline
Total      & 32 & 1,280 & 12,800 \\
\hline
\end{tabular}
\end{table}

All eight benchmark models were retrained using the same source-image partitions and evaluated on identical test observations. Their original architectures and training objectives were retained, with resource-dependent adaptations to learning rates, batch sizes, and training schedules. Benchmark training included a minimum of 100 epochs, followed by validation-based early stopping with a minimum PSNR improvement of 0.05~dB, patience of ten epochs, and a maximum of 1,000 epochs. Validation was performed every ten epochs. KAUSTics-Net was trained for 1,000 epochs, with checkpoint selection based on validation PSNR.

Table~\ref{tab:kaustics_hparams} reports the configurations and measured training durations for the completed experiments. All models were independently trained using three random seeds (1--3), and results are reported as the mean and standard deviation across seeds. Differences in architectures, GPU hardware, and training schedules mean that the reported runtimes do not constitute a matched-compute comparison. KAUSTics-Net required approximately 0.367~s per image for sampling on an NVIDIA A100 GPU, excluding image-quality metric computation. Additional model-specific implementation details are provided in the supplementary material.

\begin{table}[!h]
\centering
\caption{Training configurations and measured runtimes on KAUSTics. All models used $256 \times 256$ images. Times represent mean full-run training durations across three seeds, including validation and resume overhead.}
\label{tab:kaustics_hparams}
\footnotesize
\setlength{\tabcolsep}{3pt}
\resizebox{\columnwidth}{!}{%
\begin{tabular}{lcccc}
\hline
\textbf{Method} & \textbf{LR} & \textbf{Batch} & \textbf{GPU} & \textbf{Time (h)} \\
\hline
CycleGAN             & $2\times10^{-4}$ & 16  & A100 & 7.455 \\
UGAN                 & $1\times10^{-4}$ & 128 & V100 & 1.302 \\
FUnIE-GAN            & $3\times10^{-4}$ & 128 & V100 & 1.212 \\
DeWaterGAN           & $1\times10^{-4}$ & 8   & A100 & 21.392 \\
DiffWater            & $3\times10^{-6}$ & 1   & A100 & 20.931 \\
ResShift             & $5\times10^{-5}$ & 8   & V100 & 26.470 \\
SS-UIE               & $1\times10^{-4}$ & 9   & V100 & 8.451 \\
U-Shape Transformer  & $5\times10^{-4}$ & 8   & V100 & 6.712 \\
KAUSTics-Net (ours)  & $1\times10^{-4}$ & 12  & A100 & 4.167 \\
\hline
\end{tabular}%
}
\end{table}

\subsection{Sea-Undistort Adaptation Protocol}
\label{sec:su_adaptation}

To investigate cross-dataset transfer, KAUSTics-Net was first evaluated on Sea-Undistort without additional training. We subsequently fine-tuned the KAUSTics-trained model using the existing scene-disjoint partition of 840 training, 240 validation, and 120 test scenes. Fine-tuning used the all-effects observations, while evaluation covered the all-effects, no-waves, and no-sunglint configurations. Each distorted observation was compared against its corresponding original water-free reference image, and the same scene identifiers were maintained across configurations.

All eight benchmark methods were fine-tuned from their corresponding seed-specific, validation-selected KAUSTics checkpoints, with optimizer states reinitialized. Training used only the all-effects observations from the 840 Sea-Undistort training scenes, with the original water-free references; CycleGAN and DeWaterGAN retained their unpaired objectives using independently sampled clean images from the same training pool. Each baseline completed 100 epochs at a constant learning rate of $10^{-5}$, retaining its method-specific objective and batch size. Checkpoints were selected by mean SSIM on the 240 all-effects validation scenes, with the initial checkpoint remaining eligible. DiffWater used less frequent validation after its initial checks because of the computational cost of diffusion sampling. All methods were evaluated using three corresponding training seeds on the same 120 held-out scenes under the three distortion configurations. Complete optimization and checkpoint-selection details are provided in the supplementary material.

Fine-tuning retained the original architecture and training objective, supplemented by a sampler-aware reconstruction loss. Every fourth optimizer update included an additional loss computed from the restored output of the 25-step DDIM sampling procedure:
\begin{equation}
\mathcal{L}_{\mathrm{adapt}}
=
0.8\,\mathcal{L}_{\mathrm{RGB}}
+
0.2\,\mathcal{L}_{\mathrm{SSIM}}
+
0.2\,\mathcal{L}_{\mathrm{Lab}},
\label{eq:adapt_loss}
\end{equation}
where the RGB, SSIM, and CIELAB terms are defined in Section~\ref{sec:losses}. To limit memory consumption, gradients were propagated through only the final four sampling steps, with the preceding 21 steps detached. This provides a truncated gradient approximation of the inference procedure.

Optimization used AdamW with a learning rate of $3\times10^{-5}$, weight decay of $10^{-4}$, batch size of 12, and conditioning dropout probability of 0.1. Fine-tuning was limited to 100 epochs, with checkpoints selected according to mean validation SSIM across all 240 validation scenes. Validation was performed at epochs 0 and 1 and subsequently every five epochs. The sampling configuration remained unchanged, using strength 0.95 and classifier-free guidance scale 2.

The adaptation experiments were conducted using three independent training seeds. Since the expanded evaluation followed inspection of preliminary test results, these experiments are treated as exploratory. The zero-shot results are reported separately in the supplementary material to distinguish performance without adaptation from improvements obtained through dataset-specific fine-tuning.

\section{Results}
\label{sec:results}

\subsection{Results on the KAUSTics Dataset}
\label{sec:results_kaustics}

\subsubsection{Qualitative Comparison}

\begin{figure*}
\centering
\includegraphics[width=0.94\linewidth]{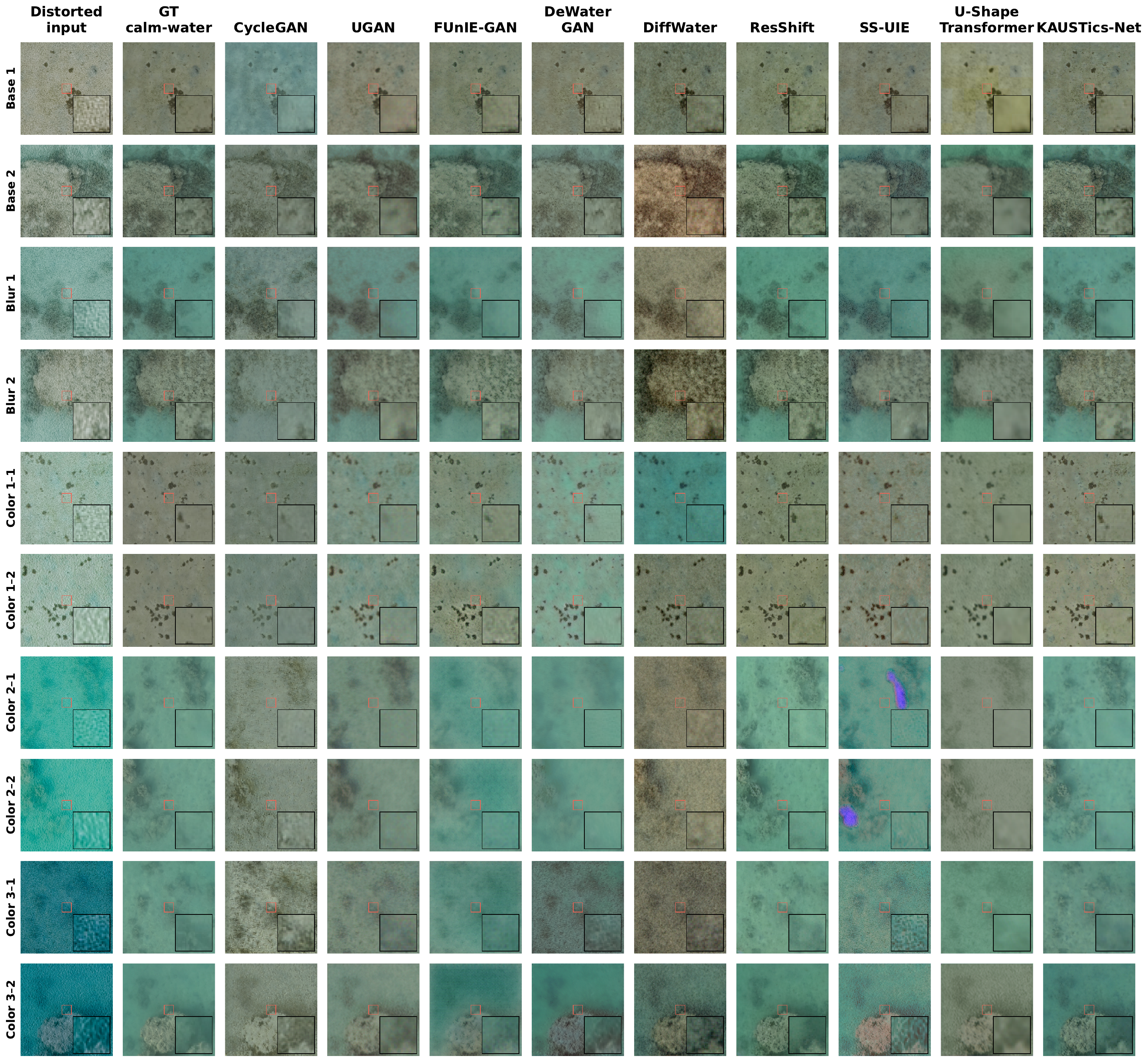}
\caption{Qualitative comparison of restoration methods on ten selected KAUSTics test patches. Rows represent the two caustic patterns under the base, blurred, and three color-transfer configurations (Color~1--3). Columns show the distorted input, calm-water reference, eight benchmark outputs, and KAUSTics-Net. All methods use identical $256 \times 256$ inputs and seed-1 checkpoints. The central zoomed regions illustrate differences in local texture and distortion suppression.}
\label{fig:methods_comparison_KAUSTICS}
\end{figure*}

Figure~\ref{fig:methods_comparison_KAUSTICS} presents a qualitative comparison of the restoration methods on ten selected patches from the held-out KAUSTics test set, covering the two caustic patterns and their corresponding blur and color-transfer variants. All methods were evaluated on identical inputs, allowing direct visual comparison of distortion suppression, color reconstruction, and preservation of seabed textures.

The GAN-based methods exhibited varying restoration characteristics. CycleGAN and UGAN reduced some distortions but retained visible differences in color and local structure relative to the reference images. FUnIE-GAN enhanced local contrast, occasionally introducing pronounced edges and high-frequency artifacts, while DeWaterGAN produced smoother appearances in several regions. Among the remaining baselines, DiffWater and ResShift recovered some structural details, although differences in texture and radiometric appearance remained. SS-UIE and U-Shape Transformer also demonstrated distinct restoration behaviors, particularly in areas containing complex seabed textures and spatially varying distortions.

KAUSTics-Net generally produced reconstructions with reduced caustic patterns and improved correspondence to the reference appearance, while retaining recognizable sandy, rocky, and vegetated structures. Differences between methods were particularly visible in regions affected by strong illumination variations and color alterations. These examples illustrate restoration behavior across the selected degradation configurations; the quantitative results provide the broader assessment over the complete held-out test set.

\subsubsection{Quantitative Comparison}

Quantitative results were evaluated on all 1,200 degraded observations from the source-image-disjoint KAUSTics test set. Tables~\ref{tab:non-ref-metrics_KAUSTics} and~\ref{tab:standard_metrics_KAUSTics} report no-reference image quality and full-reference restoration metrics, respectively, averaged across three independent training seeds.
\begin{table*}[!t]\centering
\caption{No-reference image quality metrics on the KAUSTics test set (1,200 observations). Values are reported as mean $\pm$ standard deviation across three training seeds. All methods were evaluated at $256 \times 256$ resolution.}

\label{tab:non-ref-metrics_KAUSTics}
\footnotesize
\setlength{\tabcolsep}{4pt}
\renewcommand{\arraystretch}{1.12}
\begin{adjustbox}{max width=\textwidth}
\begin{tabular}{lccccccccc}\hline
\textbf{Method} & {\bfseries\boldmath NIQE $\downarrow$} & {\bfseries\boldmath BRISQUE $\downarrow$} & {\bfseries\boldmath PIQE $\downarrow$} & {\bfseries\boldmath ILNIQE $\downarrow$} & {\bfseries\boldmath LIQE $\uparrow$} & {\bfseries\boldmath CLIPIQ $\uparrow$} & {\bfseries\boldmath MUSIQ\_AVA $\uparrow$} & {\bfseries\boldmath Sharpness} & {\bfseries\boldmath Entropy} \\ \hline
CycleGAN~\cite{zhu2017unpaired} & $\mathbf{5.870\,\pm\,0.105}$ & $15.594\,\pm\,3.547$ & $9.490\,\pm\,3.450$ & $77.654\,\pm\,3.500$ & $1.233\,\pm\,0.081$ & $0.315\,\pm\,0.062$ & $3.078\,\pm\,0.051$ & $134.265\,\pm\,60.269$ & $5.041\,\pm\,0.176$ \\
UGAN~\cite{fabbri2018enhancing} & $6.514\,\pm\,0.129$ & $27.905\,\pm\,0.448$ & $6.681\,\pm\,0.486$ & $111.196\,\pm\,2.392$ & $1.033\,\pm\,0.009$ & $0.253\,\pm\,0.010$ & $2.802\,\pm\,0.015$ & $76.006\,\pm\,8.813$ & $4.734\,\pm\,0.084$ \\
FUnIE-GAN~\cite{islam2020fast} & $7.309\,\pm\,0.555$ & $\mathbf{10.779\,\pm\,0.399}$ & $10.922\,\pm\,3.963$ & $89.018\,\pm\,7.615$ & $1.178\,\pm\,0.058$ & $0.327\,\pm\,0.008$ & $3.093\,\pm\,0.076$ & $117.203\,\pm\,33.819$ & $4.965\,\pm\,0.059$ \\
DeWaterGAN~\cite{10720144} & $8.255\,\pm\,0.736$ & $20.734\,\pm\,1.265$ & $10.867\,\pm\,4.044$ & $93.405\,\pm\,1.295$ & $1.285\,\pm\,0.051$ & $0.290\,\pm\,0.034$ & $3.037\,\pm\,0.080$ & $107.220\,\pm\,29.234$ & $4.771\,\pm\,0.080$ \\
DiffWater~\cite{guan2023diffwater} & $6.261\,\pm\,0.893$ & $23.641\,\pm\,2.762$ & $20.124\,\pm\,7.688$ & $114.717\,\pm\,1.947$ & $1.108\,\pm\,0.005$ & $0.334\,\pm\,0.009$ & $2.797\,\pm\,0.052$ & $369.749\,\pm\,157.700$ & $5.215\,\pm\,0.243$ \\
ResShift~\cite{yue2023resshift} & $11.215\,\pm\,0.079$ & $19.781\,\pm\,0.444$ & $9.695\,\pm\,1.312$ & $96.211\,\pm\,1.950$ & $\mathbf{1.568\,\pm\,0.023}$ & $0.341\,\pm\,0.006$ & $3.221\,\pm\,0.006$ & $175.156\,\pm\,14.013$ & $4.805\,\pm\,0.026$ \\
SS-UIE~\cite{peng_ssuie_2025} & $10.130\,\pm\,0.257$ & $30.025\,\pm\,3.064$ & $30.936\,\pm\,3.766$ & $\mathbf{75.313\,\pm\,5.266}$ & $1.343\,\pm\,0.059$ & $\mathbf{0.403\,\pm\,0.004}$ & $\mathbf{3.373\,\pm\,0.021}$ & $607.992\,\pm\,17.936$ & $5.085\,\pm\,0.067$ \\
U-Shape Transformer~\cite{peng2023u} & $7.626\,\pm\,0.276$ & $48.547\,\pm\,5.575$ & $39.437\,\pm\,7.657$ & $125.194\,\pm\,2.308$ & $1.127\,\pm\,0.031$ & $0.370\,\pm\,0.010$ & $2.497\,\pm\,0.044$ & $21.248\,\pm\,6.895$ & $4.421\,\pm\,0.181$ \\
KAUSTics-Net (ours) & $6.568\,\pm\,0.317$ & $16.463\,\pm\,2.072$ & $\mathbf{5.950\,\pm\,1.115}$ & $86.410\,\pm\,3.937$ & $1.364\,\pm\,0.003$ & $0.334\,\pm\,0.047$ & $3.061\,\pm\,0.045$ & $84.258\,\pm\,3.945$ & $4.662\,\pm\,0.049$ \\
\hline
\end{tabular}\end{adjustbox}
\end{table*}

\begin{table*}[!t]\centering

\caption{Full-reference restoration metrics on the KAUSTics test set (1,200 observations). Values are reported as mean $\pm$ standard deviation across three training seeds. Higher PSNR and SSIM indicate better fidelity, while lower RMSE, MSE, LPIPS, and $\Delta E_{00}$ indicate smaller reconstruction differences.}

\label{tab:standard_metrics_KAUSTics}
\footnotesize
\setlength{\tabcolsep}{4pt}
\renewcommand{\arraystretch}{1.12}
\begin{adjustbox}{max width=\textwidth}
\begin{tabular}{lcccccc}\hline
\textbf{Method} & {\bfseries\boldmath mPSNR $\uparrow$} & {\bfseries\boldmath mSSIM $\uparrow$} & {\bfseries\boldmath mRMSE $\downarrow$} & {\bfseries\boldmath mMSE $[10^{-3}]\downarrow$} & {\bfseries\boldmath LPIPS $\downarrow$} & {\bfseries\boldmath $\Delta E_{00}\downarrow$} \\ \hline
CycleGAN~\cite{zhu2017unpaired} & $24.734\,\pm\,0.923$ & $0.717\,\pm\,0.066$ & $0.065\,\pm\,0.006$ & $5.106\,\pm\,0.871$ & $0.372\,\pm\,0.055$ & $8.146\,\pm\,0.973$ \\
UGAN~\cite{fabbri2018enhancing} & $26.546\,\pm\,0.321$ & $0.777\,\pm\,0.009$ & $0.050\,\pm\,0.002$ & $2.873\,\pm\,0.225$ & $0.344\,\pm\,0.013$ & $6.479\,\pm\,0.317$ \\
FUnIE-GAN~\cite{islam2020fast} & $27.781\,\pm\,0.282$ & $0.758\,\pm\,0.020$ & $0.049\,\pm\,0.002$ & $3.290\,\pm\,0.590$ & $0.247\,\pm\,0.013$ & $5.296\,\pm\,0.311$ \\
DeWaterGAN~\cite{10720144} & $24.357\,\pm\,0.537$ & $0.769\,\pm\,0.023$ & $0.069\,\pm\,0.005$ & $5.751\,\pm\,0.939$ & $0.325\,\pm\,0.009$ & $7.492\,\pm\,0.397$ \\
DiffWater~\cite{guan2023diffwater} & $21.167\,\pm\,1.419$ & $0.605\,\pm\,0.094$ & $0.095\,\pm\,0.015$ & $10.367\,\pm\,3.349$ & $0.412\,\pm\,0.071$ & $12.091\,\pm\,2.427$ \\
ResShift~\cite{yue2023resshift} & $30.346\,\pm\,0.513$ & $0.785\,\pm\,0.008$ & $0.032\,\pm\,0.002$ & $1.140\,\pm\,0.136$ & $0.151\,\pm\,0.007$ & $4.095\,\pm\,0.192$ \\
SS-UIE~\cite{peng_ssuie_2025} & $28.387\,\pm\,0.323$ & $0.637\,\pm\,0.016$ & $0.042\,\pm\,0.001$ & $2.092\,\pm\,0.098$ & $0.438\,\pm\,0.014$ & $4.372\,\pm\,0.015$ \\
U-Shape Transformer~\cite{peng2023u} & $28.884\,\pm\,1.611$ & $0.844\,\pm\,0.016$ & $0.039\,\pm\,0.008$ & $1.755\,\pm\,0.800$ & $0.336\,\pm\,0.022$ & $4.635\,\pm\,1.378$ \\
KAUSTics-Net (ours) & $\mathbf{31.715\,\pm\,0.439}$ & $\mathbf{0.849\,\pm\,0.002}$ & $\mathbf{0.028\,\pm\,0.001}$ & $\mathbf{0.982\,\pm\,0.083}$ & $\mathbf{0.145\,\pm\,0.007}$ & $\mathbf{3.080\,\pm\,0.131}$ \\
\hline
\end{tabular}\end{adjustbox}
\end{table*}

The full-reference results in Table~\ref{tab:standard_metrics_KAUSTics} show that KAUSTics-Net achieved the highest mean PSNR of $31.715 \pm 0.439$~dB and SSIM of $0.849 \pm 0.002$. It also obtained the lowest mean RMSE ($0.028 \pm 0.001$), MSE ($0.982 \pm 0.083 \times 10^{-3}$), LPIPS ($0.145 \pm 0.007$), and $\Delta E_{00}$ ($3.080 \pm 0.131$). These results indicate improved reconstruction fidelity, structural similarity, perceptual feature agreement, and color consistency relative to the evaluated baselines.

Among the competing methods, ResShift achieved the second-highest mean PSNR ($30.346 \pm 0.513$~dB) and a competitive LPIPS of $0.151 \pm 0.007$, while U-Shape Transformer obtained the second-highest SSIM ($0.844 \pm 0.016$). SS-UIE reached $28.387 \pm 0.323$~dB PSNR but showed lower structural similarity ($0.637 \pm 0.016$), illustrating differences in reconstruction behavior among the evaluated architectures.

The no-reference results in Table~\ref{tab:non-ref-metrics_KAUSTics} show a more heterogeneous ranking. KAUSTics-Net achieved the lowest PIQE ($5.950 \pm 1.115$), while other methods obtained better scores on several naturalness-based and learned perceptual metrics. For example, FUnIE-GAN achieved the lowest BRISQUE ($10.779 \pm 0.399$), ResShift obtained the highest LIQE ($1.568 \pm 0.023$), and SS-UIE led CLIPIQ ($0.403 \pm 0.004$) and MUSIQ\_AVA ($3.373 \pm 0.021$). KAUSTics-Net therefore did not consistently outperform competing methods in predicted natural-image quality.

The differences between full-reference and no-reference metrics highlight the importance of evaluating restoration using complementary criteria. The full-reference results favor KAUSTics-Net in reconstructing the paired calm-water targets, whereas no-reference metrics produce different rankings based on their underlying image-quality assumptions. Sharpness and entropy provide additional information about edge content and intensity variability, but their values may also increase because of noise or restoration artifacts. Overall, KAUSTics-Net demonstrates strong reconstruction performance on the held-out KAUSTics data, with remaining differences between reference fidelity and predicted perceptual naturalness.

\subsection{Results on the Sea-Undistort Dataset}
\label{sec:results_su}

\subsubsection{Qualitative Comparison}

\begin{figure*}
\centering
\includegraphics[width=0.94\linewidth]{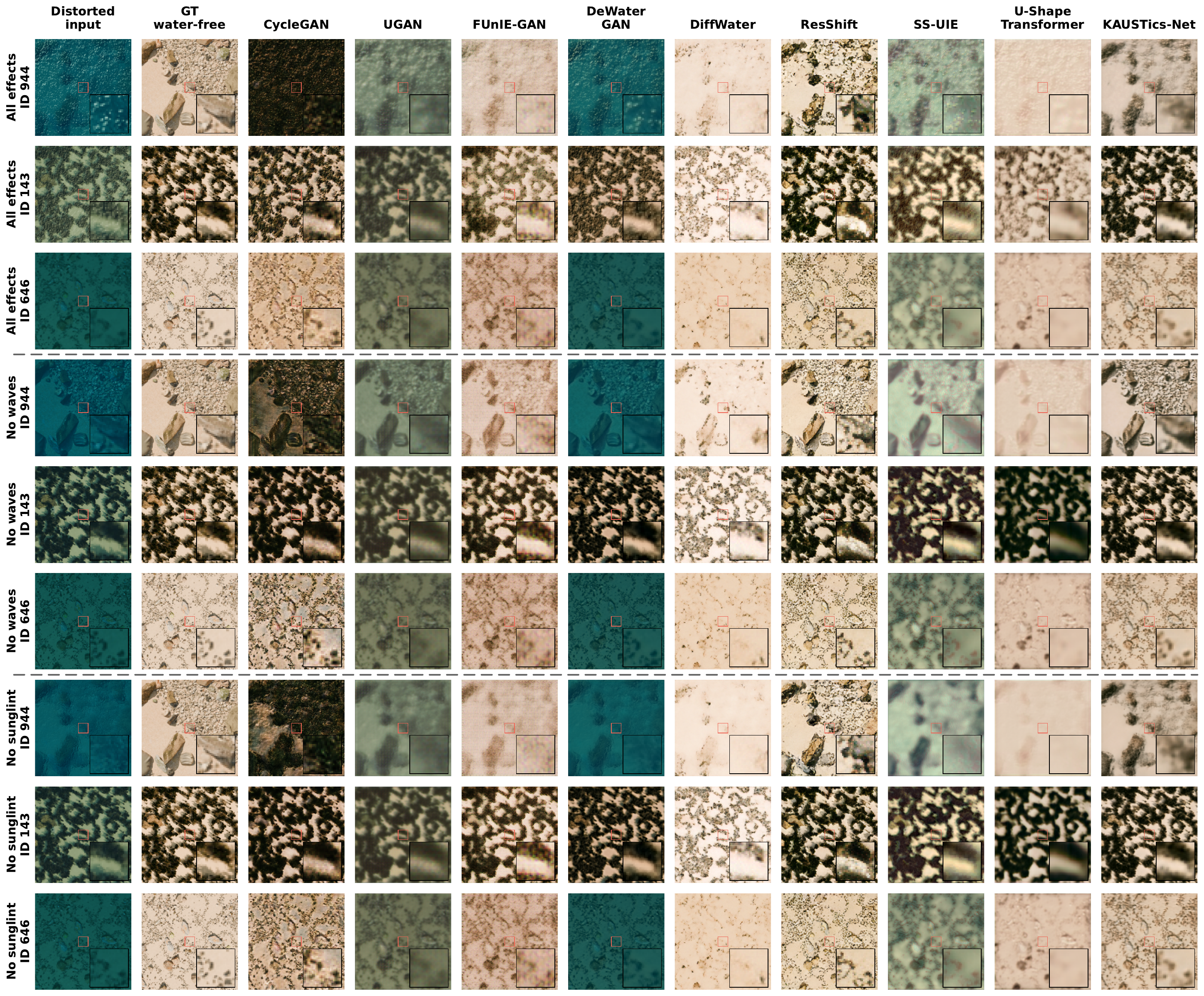}
\caption{Qualitative comparison of restoration methods on the Sea-Undistort test set. Rows show three scenes under the all-effects, no-waves, and no-sunglint configurations. Columns correspond to the distorted input, original water-free reference, eight benchmark methods, and KAUSTics-Net. All methods were evaluated at $256 \times 256$ resolution using the corresponding seed-1 checkpoints selected through validation.}
\label{fig:methods_comparison_SU}
\end{figure*}

We next evaluated the restoration methods on Sea-Undistort, which includes more severe combinations of water-surface and water-column distortions. Figure~\ref{fig:methods_comparison_SU} presents representative results from the held-out test set under the all-effects, no-waves, and no-sunglint configurations. These conditions allow the visual assessment of restoration performance under different combinations of refraction, surface reflections, and radiometric degradation.

The evaluated methods exhibited different restoration characteristics. The GAN-based models generally modified the color and contrast of distorted observations but retained visible artifacts or introduced smoothing in several regions. DiffWater, ResShift, SS-UIE, and U-Shape Transformer recovered different levels of structural detail, although reconstruction quality varied with distortion severity. Following dataset-specific adaptation, KAUSTics-Net reduced several visible surface-related artifacts and improved the overall correspondence with the reference appearance. However, some reconstructed regions remained smoother than the original references, particularly where strong distortions obscured fine seabed features. These examples illustrate the remaining challenges of recovering information under severe through-water degradation.

\subsubsection{Quantitative Comparison}

Tables~\ref{tab:non-ref-metrics_SU} and~\ref{tab:standard_metrics_SU} report the no-reference and full-reference results, respectively, across the three Sea-Undistort configurations. Each configuration contains 120 held-out scenes, and results are reported as mean and standard deviation across three independent training seeds. The original zero-shot evaluation without Sea-Undistort adaptation is reported separately in the supplementary material.

\begin{table*}[!t]\centering

\caption{Full-reference restoration metrics on Sea-Undistort under the all-effects, no-waves, and no-sunglint configurations. Each configuration comprises 120 held-out scenes, evaluated against their original water-free references. Values are reported as mean $\pm$ standard deviation across three training seeds. Higher PSNR and SSIM and lower RMSE, MSE, LPIPS, and $\Delta E_{00}$ indicate better agreement with the references.}
\label{tab:standard_metrics_SU}

\footnotesize
\setlength{\tabcolsep}{4pt}
\renewcommand{\arraystretch}{1.12}
\begin{adjustbox}{max width=\textwidth}
\begin{tabular}{lcccccc}\hline
\textbf{Method} & {\bfseries\boldmath mPSNR $\uparrow$} & {\bfseries\boldmath mSSIM $\uparrow$} & {\bfseries\boldmath mRMSE $\downarrow$} & {\bfseries\boldmath mMSE $[10^{-3}]\downarrow$} & {\bfseries\boldmath LPIPS $\downarrow$} & {\bfseries\boldmath $\Delta E_{00}\downarrow$} \\ \hline
\multicolumn{7}{l}{\textbf{All effects}} \\ \hline
Unprocessed input & $10.257$ & $0.311$ & $0.341$ & $136.826$ & $0.698$ & $32.385$ \\
CycleGAN~\cite{zhu2017unpaired} & $15.543\,\pm\,0.462$ & $0.387\,\pm\,0.019$ & $0.215\,\pm\,0.009$ & $73.756\,\pm\,4.459$ & $0.444\,\pm\,0.021$ & $16.952\,\pm\,0.978$ \\
UGAN~\cite{fabbri2018enhancing} & $10.630\,\pm\,0.122$ & $0.318\,\pm\,0.005$ & $0.305\,\pm\,0.004$ & $99.875\,\pm\,2.557$ & $0.776\,\pm\,0.013$ & $26.411\,\pm\,0.730$ \\
FUnIE-GAN~\cite{islam2020fast} & $16.886\,\pm\,0.273$ & $0.367\,\pm\,0.011$ & $0.181\,\pm\,0.006$ & $53.262\,\pm\,2.468$ & $0.497\,\pm\,0.014$ & $14.765\,\pm\,0.456$ \\
DeWaterGAN~\cite{10720144} & $11.882\,\pm\,1.088$ & $0.334\,\pm\,0.005$ & $0.318\,\pm\,0.010$ & $134.849\,\pm\,6.554$ & $0.594\,\pm\,0.051$ & $28.187\,\pm\,2.096$ \\
DiffWater~\cite{guan2023diffwater} & $10.600\,\pm\,0.948$ & $0.322\,\pm\,0.029$ & $0.383\,\pm\,0.031$ & $214.055\,\pm\,30.118$ & $0.625\,\pm\,0.041$ & $31.488\,\pm\,2.821$ \\
ResShift~\cite{yue2023resshift} & $18.022\,\pm\,0.095$ & $0.435\,\pm\,0.005$ & $0.141\,\pm\,0.000$ & $\mathbf{26.465\,\pm\,0.342}$ & $\mathbf{0.288\,\pm\,0.002}$ & $10.321\,\pm\,0.044$ \\
SS-UIE~\cite{peng_ssuie_2025} & $15.122\,\pm\,0.376$ & $0.351\,\pm\,0.017$ & $0.210\,\pm\,0.011$ & $62.568\,\pm\,7.423$ & $0.664\,\pm\,0.011$ & $18.773\,\pm\,1.196$ \\
U-Shape Transformer~\cite{peng2023u} & $15.978\,\pm\,0.179$ & $0.395\,\pm\,0.007$ & $0.203\,\pm\,0.003$ & $65.028\,\pm\,1.408$ & $0.700\,\pm\,0.007$ & $15.798\,\pm\,0.049$ \\
KAUSTics-Net (ours) & $\mathbf{19.469\,\pm\,0.332}$ & $\mathbf{0.554\,\pm\,0.005}$ & $\mathbf{0.131\,\pm\,0.011}$ & $29.598\,\pm\,6.685$ & $0.474\,\pm\,0.020$ & $\mathbf{9.865\,\pm\,1.005}$ \\
\hline
\multicolumn{7}{l}{\textbf{No waves}} \\ \hline
Unprocessed input & $10.526$ & $0.411$ & $0.342$ & $143.097$ & $0.581$ & $32.815$ \\
CycleGAN~\cite{zhu2017unpaired} & $17.277\,\pm\,0.951$ & $0.518\,\pm\,0.032$ & $0.170\,\pm\,0.020$ & $45.942\,\pm\,11.107$ & $0.324\,\pm\,0.034$ & $13.067\,\pm\,1.888$ \\
UGAN~\cite{fabbri2018enhancing} & $10.884\,\pm\,0.154$ & $0.374\,\pm\,0.009$ & $0.299\,\pm\,0.005$ & $97.485\,\pm\,3.236$ & $0.711\,\pm\,0.018$ & $25.914\,\pm\,0.792$ \\
FUnIE-GAN~\cite{islam2020fast} & $17.632\,\pm\,0.081$ & $0.437\,\pm\,0.014$ & $0.167\,\pm\,0.001$ & $47.834\,\pm\,0.466$ & $0.425\,\pm\,0.020$ & $14.066\,\pm\,0.353$ \\
DeWaterGAN~\cite{10720144} & $12.434\,\pm\,1.297$ & $0.440\,\pm\,0.006$ & $0.316\,\pm\,0.011$ & $140.994\,\pm\,9.062$ & $0.504\,\pm\,0.059$ & $28.692\,\pm\,1.938$ \\
DiffWater~\cite{guan2023diffwater} & $10.774\,\pm\,1.076$ & $0.357\,\pm\,0.035$ & $0.380\,\pm\,0.035$ & $214.568\,\pm\,33.287$ & $0.588\,\pm\,0.052$ & $31.510\,\pm\,3.061$ \\
ResShift~\cite{yue2023resshift} & $19.440\,\pm\,0.057$ & $0.546\,\pm\,0.004$ & $0.122\,\pm\,0.002$ & $21.964\,\pm\,1.215$ & $\mathbf{0.223\,\pm\,0.001}$ & $9.239\,\pm\,0.144$ \\
SS-UIE~\cite{peng_ssuie_2025} & $16.129\,\pm\,0.267$ & $0.451\,\pm\,0.014$ & $0.191\,\pm\,0.010$ & $54.635\,\pm\,7.842$ & $0.568\,\pm\,0.007$ & $17.659\,\pm\,1.314$ \\
U-Shape Transformer~\cite{peng2023u} & $17.321\,\pm\,0.245$ & $0.471\,\pm\,0.015$ & $0.173\,\pm\,0.005$ & $49.201\,\pm\,3.480$ & $0.637\,\pm\,0.016$ & $13.659\,\pm\,0.603$ \\
KAUSTics-Net (ours) & $\mathbf{20.770\,\pm\,0.051}$ & $\mathbf{0.659\,\pm\,0.006}$ & $\mathbf{0.108\,\pm\,0.002}$ & $\mathbf{18.901\,\pm\,0.807}$ & $0.370\,\pm\,0.012$ & $\mathbf{8.124\,\pm\,0.200}$ \\
\hline
\multicolumn{7}{l}{\textbf{No sunglint}} \\ \hline
Unprocessed input & $10.492$ & $0.374$ & $0.342$ & $142.779$ & $0.643$ & $32.785$ \\
CycleGAN~\cite{zhu2017unpaired} & $17.050\,\pm\,0.905$ & $0.450\,\pm\,0.028$ & $0.173\,\pm\,0.022$ & $47.030\,\pm\,13.965$ & $0.381\,\pm\,0.027$ & $13.199\,\pm\,2.153$ \\
UGAN~\cite{fabbri2018enhancing} & $10.837\,\pm\,0.145$ & $0.351\,\pm\,0.007$ & $0.301\,\pm\,0.005$ & $98.170\,\pm\,3.112$ & $0.766\,\pm\,0.014$ & $25.983\,\pm\,0.793$ \\
FUnIE-GAN~\cite{islam2020fast} & $17.444\,\pm\,0.020$ & $0.405\,\pm\,0.013$ & $0.169\,\pm\,0.000$ & $48.483\,\pm\,0.465$ & $0.478\,\pm\,0.019$ & $14.092\,\pm\,0.386$ \\
DeWaterGAN~\cite{10720144} & $12.263\,\pm\,1.180$ & $0.396\,\pm\,0.009$ & $0.318\,\pm\,0.009$ & $142.066\,\pm\,10.457$ & $0.565\,\pm\,0.059$ & $28.907\,\pm\,1.746$ \\
DiffWater~\cite{guan2023diffwater} & $10.650\,\pm\,1.002$ & $0.339\,\pm\,0.030$ & $0.384\,\pm\,0.034$ & $216.684\,\pm\,33.112$ & $0.616\,\pm\,0.043$ & $31.676\,\pm\,2.987$ \\
ResShift~\cite{yue2023resshift} & $18.414\,\pm\,0.155$ & $0.464\,\pm\,0.002$ & $0.134\,\pm\,0.004$ & $23.523\,\pm\,3.288$ & $\mathbf{0.265\,\pm\,0.002}$ & $9.842\,\pm\,0.393$ \\
SS-UIE~\cite{peng_ssuie_2025} & $15.935\,\pm\,0.289$ & $0.418\,\pm\,0.014$ & $0.194\,\pm\,0.009$ & $55.638\,\pm\,7.250$ & $0.623\,\pm\,0.003$ & $17.784\,\pm\,1.247$ \\
U-Shape Transformer~\cite{peng2023u} & $17.148\,\pm\,0.164$ & $0.449\,\pm\,0.011$ & $0.174\,\pm\,0.004$ & $49.331\,\pm\,2.541$ & $0.679\,\pm\,0.014$ & $13.657\,\pm\,0.483$ \\
KAUSTics-Net (ours) & $\mathbf{20.157\,\pm\,0.057}$ & $\mathbf{0.592\,\pm\,0.003}$ & $\mathbf{0.115\,\pm\,0.002}$ & $\mathbf{20.475\,\pm\,2.043}$ & $0.437\,\pm\,0.011$ & $\mathbf{8.477\,\pm\,0.317}$ \\
\hline
\end{tabular}\end{adjustbox}
\end{table*}

\begin{table*}[!t]\centering

\caption{No-reference image quality metrics on Sea-Undistort under the all-effects, no-waves, and no-sunglint configurations. Each configuration comprises 120 held-out scenes. Values are reported as mean $\pm$ standard deviation across three training seeds. Sharpness and entropy are descriptive indicators without performance ranking.}
\label{tab:non-ref-metrics_SU}

\footnotesize
\setlength{\tabcolsep}{4pt}
\renewcommand{\arraystretch}{1.12}
\begin{adjustbox}{max width=\textwidth}
\begin{tabular}{lccccccccc}\hline
\textbf{Method} & {\bfseries\boldmath NIQE $\downarrow$} & {\bfseries\boldmath BRISQUE $\downarrow$} & {\bfseries\boldmath PIQE $\downarrow$} & {\bfseries\boldmath ILNIQE $\downarrow$} & {\bfseries\boldmath LIQE $\uparrow$} & {\bfseries\boldmath CLIPIQ $\uparrow$} & {\bfseries\boldmath MUSIQ\_AVA $\uparrow$} & {\bfseries\boldmath Sharpness} & {\bfseries\boldmath Entropy} \\ \hline
\multicolumn{10}{l}{\textbf{All effects}} \\ \hline
Unprocessed input & $8.348$ & $30.378$ & $23.864$ & $64.283$ & $1.935$ & $0.388$ & $3.205$ & $1380.087$ & $5.211$ \\
CycleGAN~\cite{zhu2017unpaired} & $\mathbf{6.071\,\pm\,0.330}$ & $16.768\,\pm\,3.288$ & $16.382\,\pm\,2.892$ & $47.999\,\pm\,2.623$ & $1.808\,\pm\,0.152$ & $0.456\,\pm\,0.028$ & $3.155\,\pm\,0.083$ & $1414.636\,\pm\,507.040$ & $6.134\,\pm\,0.145$ \\
UGAN~\cite{fabbri2018enhancing} & $7.905\,\pm\,0.318$ & $23.939\,\pm\,4.092$ & $21.042\,\pm\,3.592$ & $100.116\,\pm\,2.558$ & $1.016\,\pm\,0.003$ & $0.168\,\pm\,0.015$ & $2.821\,\pm\,0.105$ & $136.435\,\pm\,13.936$ & $6.035\,\pm\,0.061$ \\
FUnIE-GAN~\cite{islam2020fast} & $8.394\,\pm\,0.618$ & $\mathbf{9.939\,\pm\,6.309}$ & $12.299\,\pm\,2.384$ & $74.348\,\pm\,4.000$ & $1.091\,\pm\,0.018$ & $0.310\,\pm\,0.019$ & $\mathbf{3.410\,\pm\,0.008}$ & $515.183\,\pm\,40.248$ & $6.035\,\pm\,0.064$ \\
DeWaterGAN~\cite{10720144} & $7.870\,\pm\,0.207$ & $34.232\,\pm\,0.697$ & $31.196\,\pm\,0.576$ & $67.756\,\pm\,1.008$ & $1.704\,\pm\,0.038$ & $0.325\,\pm\,0.019$ & $2.958\,\pm\,0.012$ & $937.302\,\pm\,68.339$ & $5.609\,\pm\,0.078$ \\
DiffWater~\cite{guan2023diffwater} & $8.693\,\pm\,1.522$ & $12.668\,\pm\,2.359$ & $\mathbf{5.120\,\pm\,0.439}$ & $59.549\,\pm\,13.839$ & $1.561\,\pm\,0.345$ & $0.477\,\pm\,0.066$ & $3.019\,\pm\,0.112$ & $544.385\,\pm\,187.163$ & $4.807\,\pm\,0.613$ \\
ResShift~\cite{yue2023resshift} & $9.032\,\pm\,0.018$ & $34.873\,\pm\,0.186$ & $18.645\,\pm\,0.584$ & $\mathbf{40.746\,\pm\,0.298}$ & $\mathbf{2.425\,\pm\,0.003}$ & $\mathbf{0.522\,\pm\,0.003}$ & $3.114\,\pm\,0.006$ & $3104.048\,\pm\,169.855$ & $6.320\,\pm\,0.017$ \\
SS-UIE~\cite{peng_ssuie_2025} & $7.465\,\pm\,0.626$ & $33.305\,\pm\,2.167$ & $32.425\,\pm\,2.832$ & $77.105\,\pm\,1.065$ & $1.132\,\pm\,0.043$ & $0.260\,\pm\,0.009$ & $2.830\,\pm\,0.052$ & $538.701\,\pm\,143.180$ & $5.795\,\pm\,0.047$ \\
U-Shape Transformer~\cite{peng2023u} & $10.409\,\pm\,0.176$ & $57.022\,\pm\,0.772$ & $74.144\,\pm\,3.578$ & $104.490\,\pm\,1.731$ & $1.254\,\pm\,0.012$ & $0.298\,\pm\,0.013$ & $2.593\,\pm\,0.040$ & $79.195\,\pm\,10.814$ & $5.452\,\pm\,0.140$ \\
KAUSTics-Net (ours) & $7.889\,\pm\,0.360$ & $33.707\,\pm\,1.670$ & $33.959\,\pm\,3.749$ & $79.999\,\pm\,2.717$ & $1.346\,\pm\,0.024$ & $0.352\,\pm\,0.027$ & $2.791\,\pm\,0.032$ & $165.446\,\pm\,23.531$ & $5.706\,\pm\,0.124$ \\
\hline
\multicolumn{10}{l}{\textbf{No waves}} \\ \hline
Unprocessed input & $7.769$ & $19.143$ & $18.021$ & $61.298$ & $1.690$ & $0.390$ & $3.148$ & $346.365$ & $4.682$ \\
CycleGAN~\cite{zhu2017unpaired} & $\mathbf{5.689\,\pm\,0.180}$ & $11.325\,\pm\,2.231$ & $15.419\,\pm\,2.239$ & $46.009\,\pm\,2.112$ & $1.705\,\pm\,0.135$ & $0.479\,\pm\,0.034$ & $3.235\,\pm\,0.101$ & $1193.134\,\pm\,419.752$ & $5.991\,\pm\,0.098$ \\
UGAN~\cite{fabbri2018enhancing} & $8.285\,\pm\,0.497$ & $18.569\,\pm\,4.013$ & $21.229\,\pm\,4.214$ & $95.982\,\pm\,2.725$ & $1.012\,\pm\,0.002$ & $0.160\,\pm\,0.010$ & $2.834\,\pm\,0.085$ & $138.811\,\pm\,13.137$ & $5.874\,\pm\,0.041$ \\
FUnIE-GAN~\cite{islam2020fast} & $8.582\,\pm\,0.676$ & $\mathbf{7.880\,\pm\,7.289}$ & $12.103\,\pm\,2.435$ & $72.560\,\pm\,4.970$ & $1.093\,\pm\,0.029$ & $0.311\,\pm\,0.020$ & $\mathbf{3.439\,\pm\,0.018}$ & $548.052\,\pm\,57.468$ & $6.073\,\pm\,0.101$ \\
DeWaterGAN~\cite{10720144} & $7.167\,\pm\,0.361$ & $29.635\,\pm\,2.373$ & $30.173\,\pm\,0.709$ & $65.060\,\pm\,2.043$ & $1.440\,\pm\,0.079$ & $0.312\,\pm\,0.013$ & $2.881\,\pm\,0.024$ & $288.057\,\pm\,54.979$ & $5.124\,\pm\,0.075$ \\
DiffWater~\cite{guan2023diffwater} & $8.762\,\pm\,1.567$ & $12.839\,\pm\,2.499$ & $\mathbf{5.465\,\pm\,0.641}$ & $59.643\,\pm\,13.902$ & $1.571\,\pm\,0.362$ & $0.468\,\pm\,0.064$ & $3.021\,\pm\,0.120$ & $573.531\,\pm\,173.191$ & $4.813\,\pm\,0.622$ \\
ResShift~\cite{yue2023resshift} & $9.516\,\pm\,0.134$ & $34.545\,\pm\,0.772$ & $20.079\,\pm\,0.814$ & $\mathbf{42.735\,\pm\,0.682}$ & $\mathbf{2.464\,\pm\,0.009}$ & $\mathbf{0.540\,\pm\,0.005}$ & $3.254\,\pm\,0.022$ & $3000.847\,\pm\,157.612$ & $6.097\,\pm\,0.014$ \\
SS-UIE~\cite{peng_ssuie_2025} & $6.859\,\pm\,0.763$ & $27.834\,\pm\,2.604$ & $30.564\,\pm\,2.629$ & $71.369\,\pm\,2.096$ & $1.105\,\pm\,0.025$ & $0.276\,\pm\,0.014$ & $2.955\,\pm\,0.065$ & $240.754\,\pm\,40.795$ & $5.558\,\pm\,0.032$ \\
U-Shape Transformer~\cite{peng2023u} & $11.389\,\pm\,0.261$ & $56.535\,\pm\,0.138$ & $78.882\,\pm\,3.629$ & $101.918\,\pm\,0.920$ & $1.221\,\pm\,0.024$ & $0.271\,\pm\,0.009$ & $2.637\,\pm\,0.043$ & $73.094\,\pm\,12.989$ & $5.213\,\pm\,0.134$ \\
KAUSTics-Net (ours) & $8.218\,\pm\,0.313$ & $33.144\,\pm\,1.803$ & $35.001\,\pm\,3.652$ & $71.228\,\pm\,0.976$ & $1.430\,\pm\,0.029$ & $0.363\,\pm\,0.023$ & $2.849\,\pm\,0.012$ & $282.754\,\pm\,33.265$ & $5.689\,\pm\,0.076$ \\
\hline
\multicolumn{10}{l}{\textbf{No sunglint}} \\ \hline
Unprocessed input & $6.641$ & $16.585$ & $15.804$ & $64.752$ & $1.610$ & $0.375$ & $3.082$ & $260.025$ & $4.776$ \\
CycleGAN~\cite{zhu2017unpaired} & $\mathbf{5.683\,\pm\,0.350}$ & $14.068\,\pm\,2.568$ & $15.789\,\pm\,3.195$ & $46.311\,\pm\,2.579$ & $1.746\,\pm\,0.122$ & $0.468\,\pm\,0.032$ & $3.183\,\pm\,0.105$ & $1192.952\,\pm\,463.451$ & $6.041\,\pm\,0.133$ \\
UGAN~\cite{fabbri2018enhancing} & $8.167\,\pm\,0.410$ & $18.047\,\pm\,4.308$ & $21.515\,\pm\,4.311$ & $96.712\,\pm\,3.192$ & $1.017\,\pm\,0.002$ & $0.158\,\pm\,0.013$ & $2.828\,\pm\,0.106$ & $134.458\,\pm\,13.823$ & $5.919\,\pm\,0.040$ \\
FUnIE-GAN~\cite{islam2020fast} & $8.800\,\pm\,0.699$ & $\mathbf{7.630\,\pm\,7.322}$ & $11.868\,\pm\,2.174$ & $75.268\,\pm\,4.814$ & $1.082\,\pm\,0.024$ & $0.313\,\pm\,0.024$ & $\mathbf{3.517\,\pm\,0.027}$ & $496.533\,\pm\,54.734$ & $5.984\,\pm\,0.109$ \\
DeWaterGAN~\cite{10720144} & $6.774\,\pm\,0.226$ & $28.374\,\pm\,2.417$ & $24.984\,\pm\,1.265$ & $70.394\,\pm\,1.616$ & $1.378\,\pm\,0.031$ & $0.306\,\pm\,0.008$ & $2.880\,\pm\,0.023$ & $219.488\,\pm\,43.947$ & $5.192\,\pm\,0.102$ \\
DiffWater~\cite{guan2023diffwater} & $9.007\,\pm\,1.681$ & $12.458\,\pm\,2.782$ & $\mathbf{5.188\,\pm\,0.812}$ & $60.677\,\pm\,15.710$ & $1.544\,\pm\,0.339$ & $0.476\,\pm\,0.068$ & $3.047\,\pm\,0.117$ & $519.546\,\pm\,184.063$ & $4.727\,\pm\,0.635$ \\
ResShift~\cite{yue2023resshift} & $9.100\,\pm\,0.068$ & $35.022\,\pm\,0.460$ & $19.775\,\pm\,0.367$ & $\mathbf{40.991\,\pm\,0.199}$ & $\mathbf{2.460\,\pm\,0.032}$ & $\mathbf{0.522\,\pm\,0.004}$ & $3.172\,\pm\,0.017$ & $3102.789\,\pm\,167.163$ & $6.277\,\pm\,0.016$ \\
SS-UIE~\cite{peng_ssuie_2025} & $6.819\,\pm\,0.414$ & $27.809\,\pm\,2.043$ & $29.149\,\pm\,2.639$ & $76.969\,\pm\,2.404$ & $1.100\,\pm\,0.022$ & $0.266\,\pm\,0.021$ & $2.827\,\pm\,0.049$ & $207.217\,\pm\,33.064$ & $5.535\,\pm\,0.015$ \\
U-Shape Transformer~\cite{peng2023u} & $11.023\,\pm\,0.349$ & $58.714\,\pm\,0.337$ & $78.530\,\pm\,2.636$ & $104.855\,\pm\,1.719$ & $1.290\,\pm\,0.009$ & $0.287\,\pm\,0.011$ & $2.590\,\pm\,0.046$ & $67.348\,\pm\,14.124$ & $5.159\,\pm\,0.179$ \\
KAUSTics-Net (ours) & $8.355\,\pm\,0.461$ & $34.817\,\pm\,1.725$ & $36.312\,\pm\,4.386$ & $78.138\,\pm\,1.489$ & $1.369\,\pm\,0.014$ & $0.344\,\pm\,0.023$ & $2.795\,\pm\,0.025$ & $195.470\,\pm\,29.171$ & $5.676\,\pm\,0.111$ \\
\hline
\end{tabular}\end{adjustbox}
\end{table*}

The full-reference results in Table~\ref{tab:standard_metrics_SU} show that KAUSTics-Net achieved the highest mean PSNR and SSIM across all three configurations. Under all effects, it obtained $19.469 \pm 0.332$~dB PSNR and $0.554 \pm 0.005$ SSIM, compared with $10.257$~dB and $0.311$ for the unprocessed input. Performance further improved under no waves, reaching $20.770 \pm 0.051$~dB and $0.659 \pm 0.006$, while the no-sunglint configuration yielded $20.157 \pm 0.057$~dB and $0.592 \pm 0.003$. KAUSTics-Net also achieved the lowest mean $\Delta E_{00}$ across the three configurations, indicating improved color agreement with the synthetic references.

ResShift provided the strongest competing full-reference results, reaching PSNR values of $18.022 \pm 0.095$, $19.440 \pm 0.057$, and $18.414 \pm 0.155$~dB under all effects, no waves, and no sunglint, respectively. However, ResShift consistently achieved lower LPIPS values ($0.288$, $0.223$, and $0.265$) than KAUSTics-Net ($0.474$, $0.370$, and $0.437$). These findings indicate that the improvements in pixel-wise fidelity and structural similarity were accompanied by differences in perceptual feature reconstruction.

The no-reference results in Table~\ref{tab:non-ref-metrics_SU} showed more heterogeneous rankings. Under the all-effects configuration, CycleGAN obtained the lowest NIQE ($6.071 \pm 0.330$), DiffWater the lowest PIQE ($5.120 \pm 0.439$), and ResShift the highest LIQE ($2.425 \pm 0.003$) and CLIPIQ ($0.522 \pm 0.003$). KAUSTics-Net did not consistently lead these metrics, despite its superior full-reference reconstruction scores. Similar differences were observed under the no-waves and no-sunglint conditions.

Overall, the adapted KAUSTics-Net achieved stronger agreement with the original water-free references in terms of PSNR, SSIM, and $\Delta E_{00}$ across all three configurations. The lower LPIPS values obtained by ResShift and the mixed no-reference rankings highlight remaining perceptual differences. These results demonstrate the benefits of dataset-specific adaptation while emphasizing the need to consider complementary fidelity and perceptual metrics when evaluating through-water image restoration.

\subsection{Ablation Study}
\label{sec:ablation}

To assess the contribution of individual architectural components and training objectives, we conducted an ablation study on the KAUSTics dataset. Each configuration was evaluated on the same 1,200 held-out observations across three independent training seeds. Table~\ref{tab:ablation} reports the effects of removing FiLM, cross-attention, the conditional encoder, and individual reconstruction losses, using the full KAUSTics-Net configuration as the reference.

\begin{table*}[!t]\centering
\caption{Ablation study of KAUSTics-Net on 1,200 held-out KAUSTics observations. Results are reported as mean $\pm$ standard deviation across three training seeds. Individual architectural components and loss terms are removed while retaining the remaining configuration. The component ablations use guidance scale 2, while the matched conditioning-dropout comparison uses guidance scale 1 for both configurations. Bold values indicate the best mean within each comparison block.}
\label{tab:ablation}
\footnotesize
\setlength{\tabcolsep}{4pt}
\renewcommand{\arraystretch}{1.12}
\begin{adjustbox}{max width=\textwidth}
\begin{tabular}{lcccc}\hline
\textbf{Method} & {\bfseries\boldmath mPSNR $\uparrow$} & {\bfseries\boldmath mSSIM $\uparrow$} & {\bfseries\boldmath LPIPS $\downarrow$} & {\bfseries\boldmath $\Delta E_{00}\downarrow$} \\ \hline
\multicolumn{5}{l}{\textbf{Component removals: guidance 2}} \\ \hline
Ours & $31.715\,\pm\,0.439$ & $0.849\,\pm\,0.002$ & $0.145\,\pm\,0.007$ & $3.080\,\pm\,0.131$ \\
Without FiLM & $31.089\,\pm\,0.216$ & $0.845\,\pm\,0.005$ & $0.149\,\pm\,0.003$ & $3.131\,\pm\,0.072$ \\
Without cross-attention & $31.813\,\pm\,0.760$ & $0.846\,\pm\,0.006$ & $0.146\,\pm\,0.004$ & $3.037\,\pm\,0.289$ \\
Without conditional encoder & $31.240\,\pm\,0.406$ & $0.840\,\pm\,0.013$ & $0.146\,\pm\,0.007$ & $3.128\,\pm\,0.122$ \\
Without low-frequency loss & $31.838\,\pm\,0.240$ & $0.847\,\pm\,0.006$ & $0.146\,\pm\,0.003$ & $3.007\,\pm\,0.143$ \\
Without UV loss & $\mathbf{32.181\,\pm\,0.103}$ & $0.844\,\pm\,0.009$ & $0.142\,\pm\,0.002$ & $3.048\,\pm\,0.126$ \\
Without RGB reconstruction & $32.062\,\pm\,0.234$ & $0.848\,\pm\,0.008$ & $0.148\,\pm\,0.006$ & $\mathbf{2.991\,\pm\,0.082}$ \\
Without SSIM loss & $29.994\,\pm\,0.247$ & $0.821\,\pm\,0.004$ & $0.166\,\pm\,0.004$ & $3.660\,\pm\,0.257$ \\
Without Lab loss & $31.916\,\pm\,0.610$ & $\mathbf{0.851\,\pm\,0.002}$ & $\mathbf{0.142\,\pm\,0.006}$ & $2.991\,\pm\,0.158$ \\
\hline
\multicolumn{5}{l}{\textbf{Matched dropout control: guidance 1}} \\ \hline
Ours, CFG 1 & $28.924\,\pm\,0.592$ & $0.849\,\pm\,0.004$ & $0.147\,\pm\,0.003$ & $4.143\,\pm\,0.137$ \\
Without conditioning dropout, CFG 1 & $\mathbf{30.789\,\pm\,0.260}$ & $\mathbf{0.849\,\pm\,0.001}$ & $\mathbf{0.134\,\pm\,0.001}$ & $\mathbf{3.340\,\pm\,0.063}$ \\
\hline
\end{tabular}\end{adjustbox}
\end{table*}
Among the architectural components, removing FiLM or the conditional encoder slightly reduced the mean PSNR and SSIM, indicating a modest benefit from multi-scale conditioning. Removing cross-attention produced a marginally higher PSNR (31.813 versus 31.715~dB) but lower SSIM (0.846 versus 0.849). These differences were small relative to the variability across seeds, suggesting that the contributions of individual conditioning mechanisms are not consistently reflected across all metrics.

The most pronounced degradation occurred when removing the SSIM loss, which reduced PSNR from 31.715 to 29.994~dB and SSIM from 0.849 to 0.821, while increasing LPIPS from 0.145 to 0.166 and $\Delta E_{00}$ from 3.080 to 3.660. This indicates that structural supervision plays an important role in the reconstruction objective. In contrast, removing the low-frequency, UV, RGB reconstruction, or CIELAB loss produced mixed results, with some configurations improving individual metrics. For example, removing the UV loss increased PSNR to 32.181~dB, while removing the CIELAB loss slightly improved SSIM to 0.851. These findings indicate that the complete loss formulation does not optimize every metric individually.

Conditioning dropout was evaluated separately using a matched classifier-free guidance scale of 1 for both configurations. Removing dropout preserved the mean SSIM (0.849) while improving PSNR, LPIPS, and $\Delta E_{00}$. Thus, the ablation does not establish an independent advantage from conditioning dropout under this sampling configuration.

Overall, the ablation results identify the SSIM loss as the component with the clearest contribution to reconstruction performance. The remaining architectural and loss components exhibit smaller or metric-dependent effects, highlighting the importance of assessing the proposed configuration across multiple restoration criteria.

\subsection{Real-UAV Downstream Evaluation}
\label{sec:real_uav}

To examine the behavior of the restoration methods on real through-water imagery, we conducted an exploratory case study using a UAV frame from the controlled shallow-water survey of Scilla et al.~\cite{scilla2025computer}. The comparison reference was generated using the temporal computer-vision correction approach developed in that study, which combines information from 60 consecutive frames. This image provides a low-distortion visual reference but does not constitute independently validated water-free ground truth. Two regions containing a ground control point (GCP) and a bucket, located at approximately 0.600 and 0.900~m water depth, respectively, were selected for evaluation.

All restoration methods were applied to identical $1000 \times 1000$~pixel image contexts resized to $256 \times 256$ pixels, using their Sea-Undistort-adapted seed-1 checkpoints without further training. To reduce differences in large-scale color appearance, the model outputs received the same low-frequency color harmonization based on the distorted input, using a Gaussian smoothing scale of $\sigma=5$ pixels and correction strength $\alpha=1$. The input and temporal reference remained unchanged, and no geometric registration was applied. This processing was specific to the real-UAV case study and was not used in the quantitative benchmark evaluations.

\begin{figure*}
\centering
\includegraphics[width=0.94\linewidth]{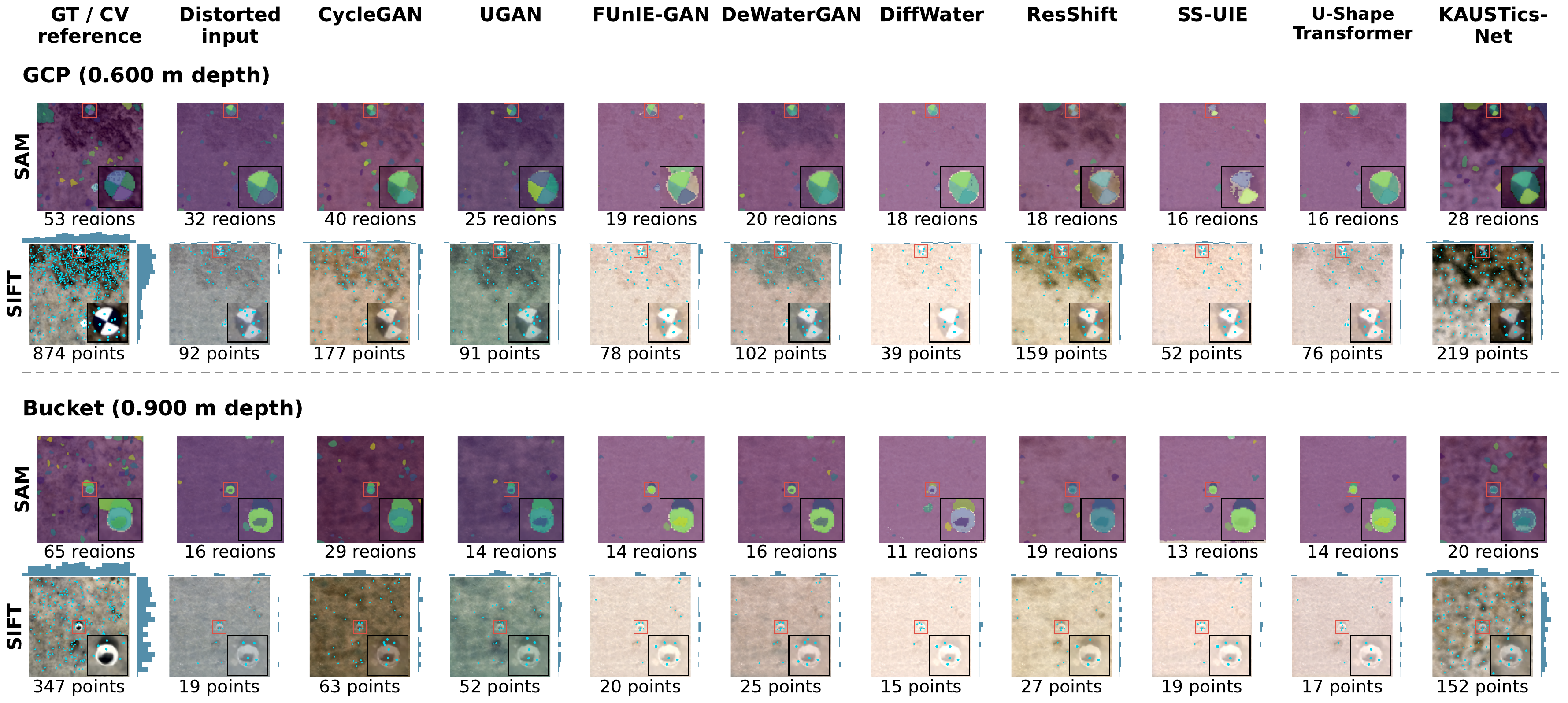}
\caption{Real-UAV structural diagnostics for the GCP (0.600~m depth) and bucket (0.900~m depth). Columns show the temporal computer-vision reference, distorted input, and eight benchmark restorations alongside KAUSTics-Net. Each object is evaluated using SAM segmentation regions and SIFT keypoints. All model outputs receive identical low-frequency color harmonization ($\alpha=1$, $\sigma=5$ pixels), while the input and reference remain unchanged. Red boxes identify corresponding object-centered regions, and insets enlarge these regions for visualization. Histograms show the spatial distribution of SIFT keypoints using 20 bins and a common count scale within each object. SAM colors distinguish automatically generated regions and do not represent semantic classes. All methods use identical SAM and SIFT settings.}
\label{fig:real_uav_diagnostics}
\end{figure*}

Figure~\ref{fig:real_uav_diagnostics} compares automatic segmentation regions produced by the Segment Anything Model (SAM) and local features detected using the scale-invariant feature transform (SIFT). SAM ViT-B was applied with a $32 \times 32$ point grid, predicted-IoU threshold of 0.88, and stability threshold of 0.95. SIFT used a contrast threshold of 0.040, with identical detector settings across all images. These diagnostics characterize segmentation behavior and local feature detection without requiring manually annotated object masks.

For the GCP region, SIFT detected 92 keypoints in the distorted input, 219 in the KAUSTics-Net reconstruction, and 874 in the temporal reference. The corresponding counts for the bucket were 19, 152, and 347. Although KAUSTics-Net increased the number of detected features relative to the distorted input, substantial differences remained compared with the reference. Other restoration methods exhibited different feature densities, demonstrating that the effect of image restoration on keypoint detection depends on the model and scene content.

\begin{figure}
\centering
\includegraphics[width=0.88\linewidth]{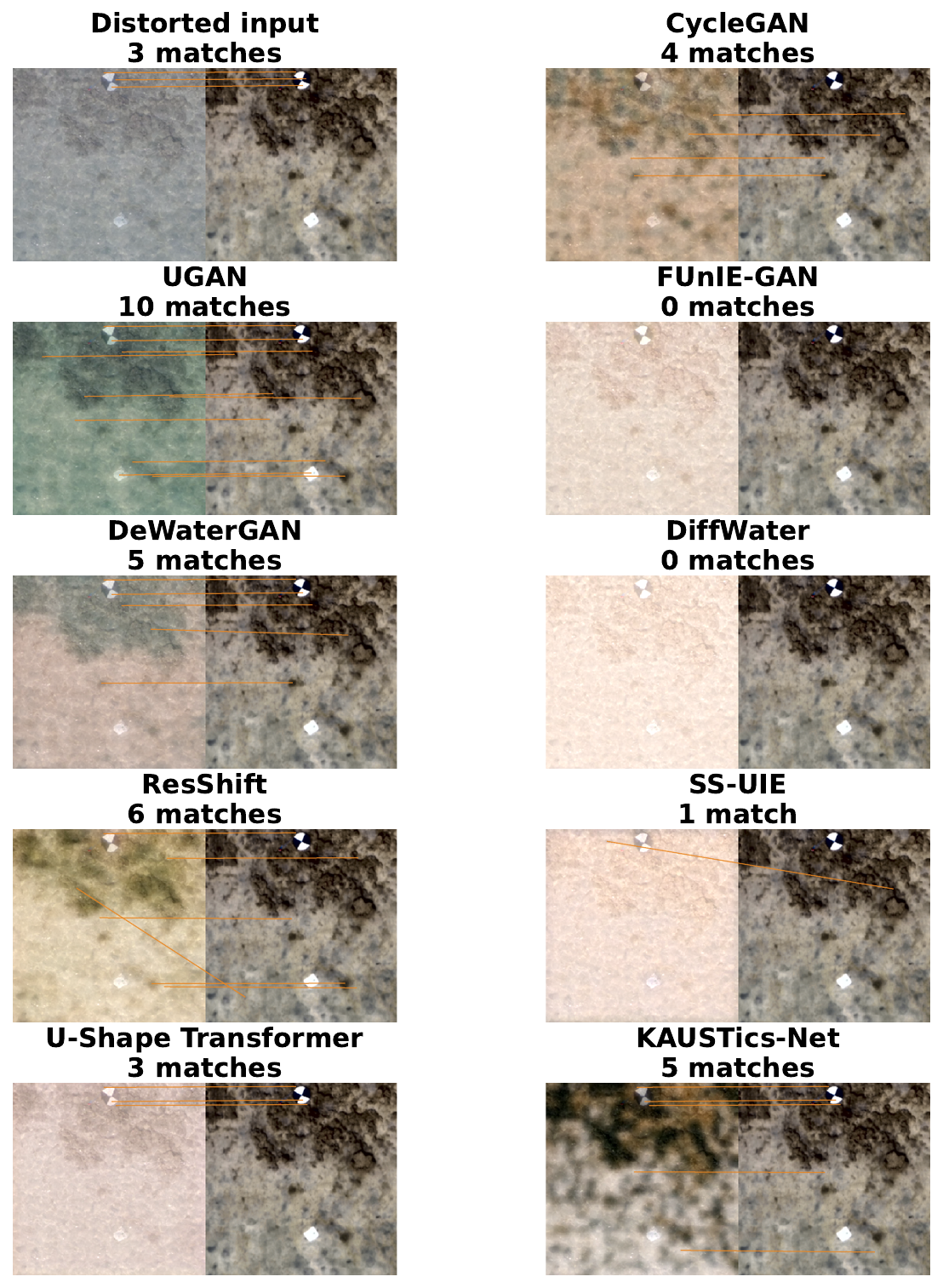}

\caption{SIFT feature correspondence between the distorted input or restored GCP images and the temporal computer-vision reference. Each panel shows the evaluated image on the left and the common reference on the right. Orange lines indicate mutual nearest-neighbor descriptor matches obtained using a bidirectional ratio threshold of 0.75. SIFT features were extracted with a contrast threshold of 0.040. Match counts represent putative correspondences without RANSAC-based geometric verification. No spatial registration was applied to the restored images.}
\label{fig:real_uav_sift}

\end{figure}

To further examine feature correspondence, Fig.~\ref{fig:real_uav_sift} compares SIFT descriptor matches between the restored images and the temporal reference. Using mutual nearest-neighbor matching with a bidirectional ratio threshold of 0.75, KAUSTics-Net produced five putative matches for the GCP, compared with three for the distorted input, and one for the bucket, compared with none for the input. UGAN produced ten matches in each region, while DiffWater produced no matches despite detecting local features. These results indicate that increased keypoint density does not necessarily correspond to increased descriptor agreement with the reference.

The SAM results also varied among methods, with changes in the number and spatial distribution of automatically generated regions (Fig.~\ref{fig:real_uav_diagnostics}). However, these regions do not represent validated semantic classes, and no independent segmentation annotations were available to quantify accuracy. Similarly, the SIFT correspondences were not geometrically verified using RANSAC or an equivalent consistency test. Consequently, the observed changes cannot establish improvements in object localization, geometric reconstruction, or habitat mapping accuracy.

Overall, this case study illustrates how restoration can alter the outputs of subsequent computer-vision operations on real UAV imagery. The results are descriptive and limited to two selected regions from one survey frame. A broader evaluation with independent geometric measurements and annotated seabed features is required to determine whether these changes translate into improved remote-sensing products.

\section{Discussion}
\label{sec:discussion}

\subsection{Generalization and Limitations}

Obtaining paired distorted and reference UAV imagery of the same shallow-water scene remains challenging because water-surface conditions can change rapidly, while repeated acquisitions under comparable illumination, tide, and viewing geometry are difficult to achieve. Consequently, learning-based restoration methods frequently rely on synthetic data, unpaired training, or domain adaptation \cite{PURNIMA2024110723,barbosa2025physically,wang2023domain,jiang2022two}. KAUSTics addresses this data limitation by combining real UAV imagery with rendered caustics and refraction, complemented by empirical color-transfer and blur augmentations. This approach enables controlled restoration experiments while retaining the textures and spatial characteristics of real shallow-water UAV observations \cite{agrafiotis2020correcting,scilla2025computer}.

The revised evaluation demonstrates improved reconstruction fidelity on held-out KAUSTics images and substantial gains following adaptation to Sea-Undistort. However, these findings have distinct implications for generalization. The KAUSTics test set contains 1,200 degraded observations derived from only three source UAV images, and the source-image-disjoint partition does not guarantee geographic independence between potentially overlapping survey footprints. Similarly, the Sea-Undistort experiments evaluate one independently generated synthetic dataset, with the main improvements obtained after dataset-specific fine-tuning. These results establish performance under the evaluated conditions but do not demonstrate reliable zero-shot transfer to arbitrary coastal environments, sensors, or acquisition geometries.

Additional limitations arise from the dataset construction. The KAUSTics reference images were acquired through water under relatively calm conditions and may retain residual geometric and radiometric distortions. Furthermore, the Blender simulations use planar seabed textures and selected water-surface configurations, while blur and color transfer provide empirical approximations of water-column degradation. The resulting observations do not reproduce the complete interaction between three-dimensional bathymetry, water constituents, multiple scattering, and sensor radiometry. Consequently, reconstruction accuracy is measured relative to the available reference appearance, and the recovery of absolute seabed geometry or intrinsic reflectance remains unverified.

Severe distortions introduce a further limitation because sun glint, scattering, and wave-induced displacement can obscure information needed for faithful reconstruction \cite{kromer2025sea}. Learning-based models may generate plausible structures in regions where the observations provide insufficient evidence, potentially reflecting correlations learned from the training distribution \cite{ye2024spurious,mulchandani2025severing}. This risk reinforces the importance of evaluating restored imagery against independent geometric and ecological measurements before using it for quantitative remote-sensing applications.

\subsection{Perceptual Metrics, Pixel Fidelity, and Task Utility}

The revised experiments highlight differences between reconstruction fidelity and predicted perceptual quality. On KAUSTics, KAUSTics-Net achieved the highest mean PSNR and SSIM and the lowest LPIPS and $\Delta E_{00}$ among the evaluated methods. Following Sea-Undistort adaptation, it also obtained the highest PSNR and SSIM across all three distortion configurations. However, the no-reference rankings remained heterogeneous, and ResShift consistently achieved lower LPIPS on Sea-Undistort. These findings indicate that improvements in pixel-level similarity, perceptual feature agreement, and natural-image statistics do not necessarily coincide, consistent with limitations reported in underwater image quality assessment \cite{li2023underwater}.

The component ablations further clarify these results. Removing the SSIM loss produced the most pronounced degradation across the evaluated fidelity and perceptual metrics, supporting its contribution to the reconstruction objective. In contrast, removing individual conditioning mechanisms or other loss terms produced smaller and sometimes favorable changes in specific metrics. The complete architecture therefore represents one experimentally evaluated configuration, without establishing that every component independently contributes to all aspects of restoration quality.

For shallow-water remote sensing, improved image quality is valuable only when relevant scene information is preserved. Accurate feature correspondence, object localization, habitat interpretation, and bathymetric reconstruction depend on geometric consistency and reliable seabed information \cite{agrafiotis2020correcting,scilla2025computer}. The real-UAV case study illustrates this distinction: restoration altered SIFT keypoint density and SAM segmentation behavior, but the number of repeatable descriptor matches with the temporal computer-vision reference remained limited. Moreover, the SAM regions lacked independent semantic annotations, and the feature matches were not geometrically verified. The results therefore demonstrate changes in downstream computer-vision outputs without establishing improvements in mapping accuracy. Future task-oriented evaluations should incorporate annotated habitat boundaries, independently surveyed features, or validated bathymetric measurements.

\subsection{Practical Considerations and Future Directions}

KAUSTics-Net integrates established conditional diffusion components within a restoration framework designed for through-water UAV imagery. Its contribution lies in the combination of semi-synthetic data construction, multi-scale conditioning, and a composite reconstruction objective evaluated under controlled degradation settings. The ablation study provides partial support for these design choices, with the clearest benefit associated with structural supervision. Additional experiments are required to isolate the effects of individual conditioning mechanisms and determine their contributions under different training conditions.

Computational requirements also affect practical applicability. Although all methods were evaluated at a common $256 \times 256$ resolution, differences in training schedules, optimization procedures, and GPU hardware prevent a strictly matched-compute comparison. KAUSTics-Net required approximately 0.367~s per image for inference on an NVIDIA A100 GPU, excluding metric computation. Its iterative sampling procedure introduces additional considerations for processing large UAV surveys, particularly when preserving the native spatial resolution of high-resolution acquisitions. Comparisons under matched computational budgets and evaluation at larger image resolutions would help establish the trade-offs between restoration quality and operational efficiency.

Future research should prioritize spatially independent UAV surveys covering additional water types, depths, illumination conditions, and sensors, together with physically informed simulations incorporating measured water-column properties and seabed geometry. Limited real-data adaptation and uncertainty-aware restoration may further help address domain shifts and regions where seabed information is severely obscured \cite{wang2023domain,jiang2022two}. Ultimately, independently validated downstream assessments are needed to establish whether the observed restoration improvements support more accurate coastal mapping and environmental monitoring.

\section{Conclusion}
\label{sec:conclusion}

This study introduced KAUSTics, a semi-synthetic dataset comprising 1,280 reference patches derived from 32 shallow-water UAV images and 12,800 paired degraded observations, together with KAUSTics-Net, a pixel-space conditional diffusion framework for through-water image restoration. The framework integrates multi-scale conditioning, FiLM-modulated residual blocks, cross-attention, and a composite training objective combining velocity prediction with structural and chromatic reconstruction losses.

The revised evaluation used source-image-disjoint partitions, a common $256 \times 256$ image resolution, eight benchmark methods, and three independent training seeds. On KAUSTics, KAUSTics-Net achieved $31.715 \pm 0.439$~dB PSNR, $0.849 \pm 0.002$ SSIM, and $0.145 \pm 0.007$ LPIPS. Following dataset-specific adaptation, it also achieved the highest PSNR and SSIM across the three evaluated Sea-Undistort configurations, although ResShift obtained better LPIPS and several no-reference metrics favored competing methods. Architectural ablations identified structural supervision as a particularly important component, while the real-UAV case study documented changes in feature detection and segmentation behavior.

The principal limitations concern the small number of independent source scenes, the synthetic-to-real domain gap, the absence of independently validated water-free references for KAUSTics, and the limited evidence of downstream mapping improvements. Further validation across geographically independent UAV surveys, measured water-column conditions, and quantitative remote-sensing tasks is required. Within these constraints, the study demonstrates the potential of semi-synthetic data and conditional diffusion for restoring shallow-water UAV imagery toward low-distortion reference appearances.

\section*{Data and Code Availability}

The KAUSTics dataset, pretrained KAUSTics-Net models, and source code will be made publicly available upon publication at \url{https://huggingface.co/datasets/dar96io/KAUSTics}, \url{https://huggingface.co/dar96io/KAUSTics-Net}, and \url{https://github.com/Dark3850/KAUSTics-Net}, respectively. The fixed Sea-Undistort experimental partition will be available at \url{https://huggingface.co/datasets/dar96io/Sea-Undistort-evaluation}, retaining the original dataset's attribution and CC BY-NC-SA 4.0 license.

\ifCLASSOPTIONcaptionsoff
  \newpage
\fi

\phantomsection
\addcontentsline{toc}{section}{References}
\bibliographystyle{IEEEtran}
\bibliography{bibliography}

\clearpage
\appendices
\clearpage
\begingroup
\renewcommand{\theHsection}{supp.\arabic{section}}
\renewcommand{\theHsubsection}{\theHsection.\arabic{subsection}}
\renewcommand{\theHtable}{supp.\arabic{table}}
\renewcommand{\theHfigure}{supp.\arabic{figure}}
\setcounter{section}{0}
\setcounter{subsection}{0}
\setcounter{table}{0}
\setcounter{figure}{0}
\renewcommand{\thesection}{S\Roman{section}}
\renewcommand{\thesubsection}{\thesection.\Alph{subsection}}
\renewcommand{\thetable}{S\arabic{table}}
\renewcommand{\thefigure}{S\arabic{figure}}
\section*{Supplementary Material}
This supplement documents the sampler-aware adaptation procedure, provides full zero-shot Sea-Undistort comparisons, and illustrates the reference-patch inventory and appearance variation in KAUSTics. The zero-shot and fine-tuned evaluations are distinct. Sea-Undistort full-reference metrics are computed against the original synthetic water-free targets on the same 120 held-out scene identifiers per input configuration at $256\times256$ RGB resolution.

\section{Additional Adaptation Implementation Details}
Sampler-aware fine-tuning was performed in FP32 on one NVIDIA A100 GPU per seed, with gradient clipping at a global norm of 1. Each epoch consisted of 70 optimizer updates. The extra sampled-output objective was evaluated on updates 0, 4, $\ldots$, 68 (18 of 70 updates), using microbatches of four to cover the effective batch of 12. The sampling procedure ran in evaluation mode. Gradients from the standard training objective and the sampler-aware objective were combined before each optimizer step. Sampled RGB mean squared error (MSE) used unclipped outputs, while the SSIM and blurred-CIELAB terms used the clipped RGB range. The first 21 DDIM sampling steps were detached, and differentiation was restricted to the last four steps.
Early stopping required at least 60 epochs and used a patience of five validation checks, with a minimum SSIM improvement of $10^{-4}$. Each seed was initialized from its corresponding KAUSTics checkpoint, with a newly initialized AdamW optimizer. All three runs reached 100 epochs; validation-selected checkpoints were obtained at epochs 95, 100, and 100. Selection used the mean SSIM for all-effects validation scenes; the no-waves validation results were secondary. Test scene identifiers were excluded from optimization and checkpoint selection. However, preliminary Sea-Undistort test results had been inspected during the revision process, so the expanded adaptation experiment is exploratory and is not a newly blind external validation.
An initial LPIPS guard permitting an increase of at most 0.02 was not satisfied by any seed. This criterion was not retrospectively relaxed in the experimental record. Differences in model-specific schedules and the additional KAUSTics-Net adaptation experiments also mean that training and tuning effort were not matched across methods.

\subsection{Sea-Undistort Fine-Tuning of the Benchmark Methods}
\label{sec:supp_baseline_adaptation}

All eight baselines: CycleGAN, UGAN, FUnIE-GAN, DeWaterGAN, DiffWater, ResShift, SS-UIE, and U-Shape Transformer, were initialized from their corresponding seed-specific, validation-selected KAUSTics-trained restoration checkpoints. These were the same checkpoints used for the zero-shot Sea-Undistort evaluation. Restoration networks were therefore adapted from KAUSTics, rather than trained from scratch on Sea-Undistort. Optimizer states were reinitialized when adaptation began.

Every baseline used the same fixed partition of 840 training, 240 validation, and 120 test scene IDs, at RGB $256\times256$ resolution. Training used only the all-effects distorted inputs and the original water-free reference images from the 840 training scenes. No no-waves or no-sunglint configurations were added to the baseline training sets. CycleGAN and DeWaterGAN retained their unpaired training formulations: clean targets were sampled independently from the same 840-scene training pool. The other methods used the corresponding paired references.

All baselines completed 100 fine-tuning epochs with a constant learning rate of $10^{-5}$ and no early stopping. Each epoch traversed all 840 distorted training inputs. Method-specific optimization objectives were retained; U-Shape Transformer continued its final L1-based training stage. Batch sizes were 8 for CycleGAN, DeWaterGAN, ResShift, and U-Shape Transformer; 128 for UGAN and FUnIE-GAN; 9 for SS-UIE; and 1 for DiffWater. 

Checkpoint selection used the highest mean SSIM over all 240 all-effects validation scenes. The initial KAUSTics checkpoint, evaluated at epoch~0, remained eligible. For seven baselines, validation was performed at epochs 0, 1, 5, and every five epochs thereafter. For DiffWater, validation checks were performed every 20 epochs to reduce the cost of its native diffusion sampler. Consequently, validation frequency, and therefore checkpoint-selection opportunities, were not identical across methods.

Every baseline was fine-tuned using seeds 1, 2, and 3, starting from the matching KAUSTics training seed. The 120 test scene IDs were excluded from optimization and per-run checkpoint selection. Each selected checkpoint was evaluated on all three test-input variants, all-effects, no-waves, and no-sunglint, against the same original water-free reference.

\section{Zero-Shot Sea-Undistort Evaluation}
\subsection{Protocol and Full-Reference Results}
Zero-shot transfer was evaluated using the KAUSTics-trained checkpoints without any Sea-Undistort optimization. This term refers to the source-trained experimental checkpoints, not externally released model weights. All eight benchmark methods used their completed KAUSTics training runs under the 100-epoch warm-up and patience-10 rule, with validation-selected checkpoints. KAUSTics-Net used its longer-trained, validation-selected checkpoint. These training schedules were not matched for total computation. The comparison uses identical held-out scenes under all-effects, no-waves, and no-sunglint inputs, each evaluated against the original water-free reference image.

On all-effects observations, KAUSTics-Net achieved $9.418\pm0.094$~dB PSNR and $0.233\pm0.012$ SSIM, below the unprocessed input ($10.257$~dB and $0.311$, respectively). The corresponding zero-shot scores were also limited under no waves ($9.652\pm0.139$~dB, $0.291\pm0.022$) and no sunglint ($9.632\pm0.147$~dB, $0.268\pm0.019$). These results document a substantial synthetic-domain shift and should not be conflated with the gains achieved after Sea-Undistort fine-tuning in the main manuscript. Table~\ref{tab:supp_zeroshot_full} presents the complete full-reference comparison.

\begin{figure*}
\centering
\includegraphics[width=0.94\linewidth]{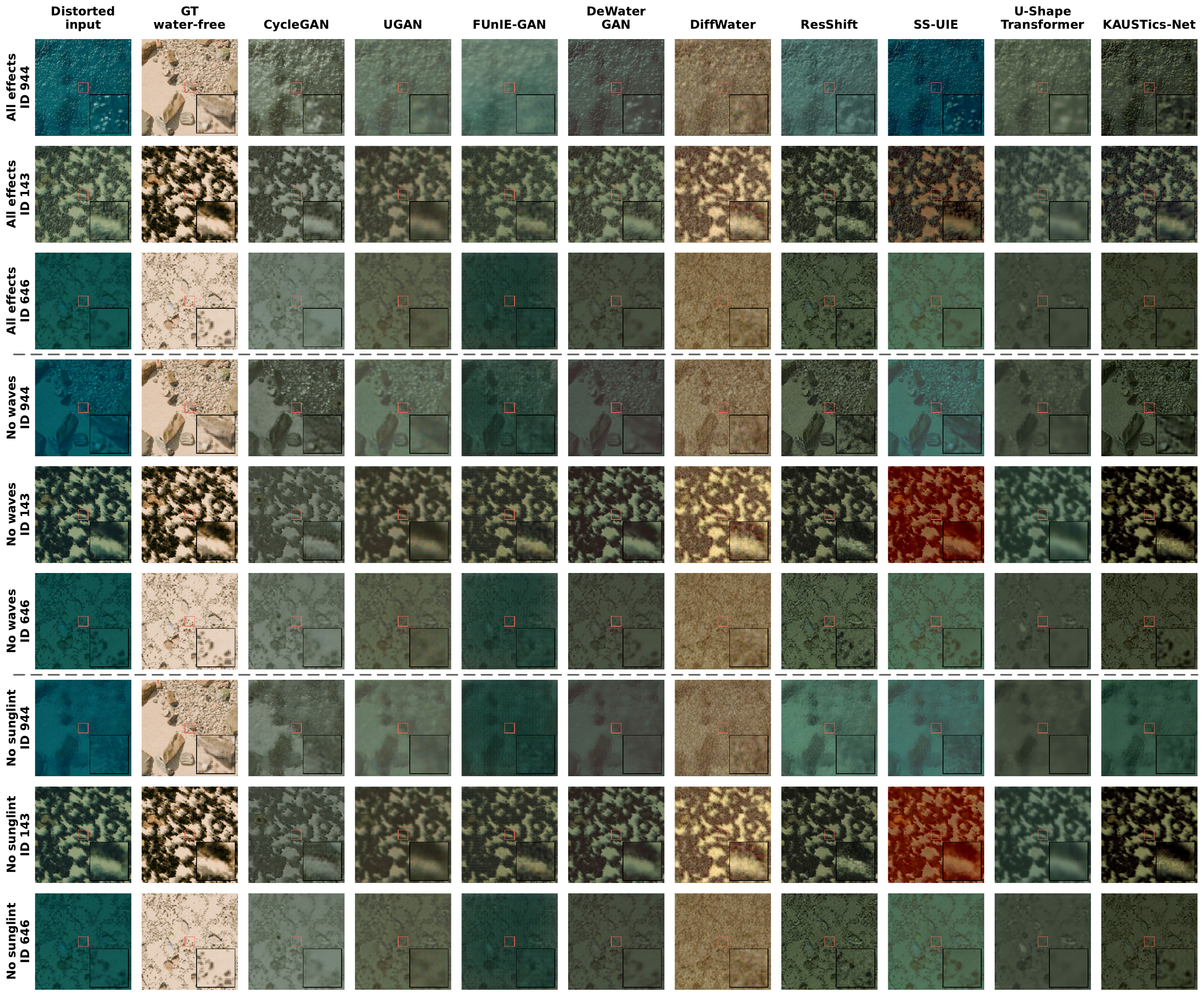}
\caption{Sea-Undistort zero-shot comparison using seed-1 checkpoints at 256 x 256 RGB resolution. Row groups show all effects, no waves, and no sunglint; each group contains scenes 944, 143, and 646 in that order. Columns show the distorted input, original water-free reference, and eight benchmark methods alongside KAUSTics-Net. Each panel uses the same central 26 x 26 pixel inset. No panel-specific colour correction was applied. All methods use the completed KAUSTics-trained checkpoints from Tables S1 and S2.}
\end{figure*}

\begin{table*}[!t]
\centering
\caption{Sea-Undistort zero-shot full-reference metrics on 120 held-out scenes per configuration. Results are mean $\pm$ sample standard deviation over three source-training seeds; all results are evaluated against the original water-free reference. MSE is shown in units of $10^{-3}$, and RMSE uses $[0,1]$ RGB units. The unprocessed input is provided for context.}
\label{tab:supp_zeroshot_full}
\footnotesize
\setlength{\tabcolsep}{3pt}
\renewcommand{\arraystretch}{1.12}
\resizebox{\textwidth}{!}{%
\begin{tabular}{lcccccc}
\hline
\textbf{Method} & {\bfseries\boldmath mPSNR $\uparrow$} & {\bfseries\boldmath mSSIM $\uparrow$} & {\bfseries\boldmath mRMSE $\downarrow$} & {\bfseries\boldmath mMSE $[10^{-3}]\downarrow$} & {\bfseries\boldmath LPIPS $\downarrow$} & {\bfseries\boldmath $\Delta E_{00}\downarrow$} \\
\hline
\multicolumn{7}{l}{\textbf{All effects}} \\
\hline
Unprocessed input & $10.257$ & $0.311$ & $0.341$ & $136.826$ & $0.698$ & $32.385$ \\
CycleGAN~\cite{zhu2017unpaired} & $\mathbf{10.157 \pm 0.064}$ & $\mathbf{0.324 \pm 0.007}$ & $\mathbf{0.317 \pm 0.002}$ & $\mathbf{105.059 \pm 1.236}$ & $0.714 \pm 0.029$ & $28.996 \pm 0.282$ \\
UGAN~\cite{fabbri2018enhancing} & $10.140 \pm 0.095$ & $0.305 \pm 0.005$ & $0.325 \pm 0.003$ & $114.512 \pm 2.043$ & $0.771 \pm 0.014$ & $\mathbf{28.376 \pm 0.517}$ \\
FUnIE-GAN~\cite{islam2020fast} & $9.249 \pm 0.156$ & $0.204 \pm 0.011$ & $0.369 \pm 0.015$ & $152.392 \pm 16.087$ & $0.744 \pm 0.011$ & $34.011 \pm 1.487$ \\
DeWaterGAN~\cite{10720144} & $10.081 \pm 0.185$ & $0.309 \pm 0.009$ & $0.343 \pm 0.009$ & $134.131 \pm 7.198$ & $0.719 \pm 0.028$ & $31.500 \pm 1.046$ \\
DiffWater~\cite{guan2023diffwater} & $9.800 \pm 0.286$ & $0.227 \pm 0.032$ & $0.334 \pm 0.009$ & $118.417 \pm 5.690$ & $0.745 \pm 0.052$ & $29.758 \pm 2.668$ \\
ResShift~\cite{yue2023resshift} & $9.639 \pm 0.208$ & $0.253 \pm 0.013$ & $0.355 \pm 0.009$ & $141.497 \pm 8.180$ & $\mathbf{0.664 \pm 0.013}$ & $31.961 \pm 0.702$ \\
SS-UIE~\cite{peng_ssuie_2025} & $9.750 \pm 0.127$ & $0.288 \pm 0.000$ & $0.356 \pm 0.009$ & $145.141 \pm 7.928$ & $0.698 \pm 0.012$ & $32.213 \pm 1.406$ \\
U-Shape Transformer~\cite{peng2023u} & $9.082 \pm 0.065$ & $0.250 \pm 0.009$ & $0.379 \pm 0.004$ & $160.612 \pm 4.400$ & $0.841 \pm 0.009$ & $35.083 \pm 1.073$ \\
KAUSTics-Net (ours) & $9.418 \pm 0.094$ & $0.233 \pm 0.012$ & $0.373 \pm 0.004$ & $160.532 \pm 6.328$ & $0.732 \pm 0.054$ & $33.632 \pm 0.053$ \\
\hline
\multicolumn{7}{l}{\textbf{No waves}} \\
\hline
Unprocessed input & $10.526$ & $0.411$ & $0.342$ & $143.097$ & $0.581$ & $32.815$ \\
CycleGAN~\cite{zhu2017unpaired} & $10.091 \pm 0.257$ & $\mathbf{0.401 \pm 0.011}$ & $\mathbf{0.320 \pm 0.010}$ & $\mathbf{106.660 \pm 6.588}$ & $0.633 \pm 0.028$ & $29.302 \pm 0.524$ \\
UGAN~\cite{fabbri2018enhancing} & $\mathbf{10.343 \pm 0.134}$ & $0.353 \pm 0.007$ & $0.322 \pm 0.004$ & $113.909 \pm 2.694$ & $0.707 \pm 0.019$ & $\mathbf{28.064 \pm 0.567}$ \\
FUnIE-GAN~\cite{islam2020fast} & $9.581 \pm 0.183$ & $0.238 \pm 0.010$ & $0.364 \pm 0.019$ & $153.043 \pm 20.183$ & $0.697 \pm 0.007$ & $33.737 \pm 1.863$ \\
DeWaterGAN~\cite{10720144} & $10.313 \pm 0.250$ & $0.382 \pm 0.018$ & $0.343 \pm 0.010$ & $138.365 \pm 7.541$ & $0.640 \pm 0.051$ & $31.802 \pm 1.098$ \\
DiffWater~\cite{guan2023diffwater} & $9.862 \pm 0.339$ & $0.257 \pm 0.033$ & $0.332 \pm 0.011$ & $117.030 \pm 7.235$ & $0.722 \pm 0.055$ & $29.652 \pm 3.063$ \\
ResShift~\cite{yue2023resshift} & $9.844 \pm 0.335$ & $0.313 \pm 0.017$ & $0.353 \pm 0.012$ & $142.829 \pm 10.533$ & $0.601 \pm 0.024$ & $31.798 \pm 1.040$ \\
SS-UIE~\cite{peng_ssuie_2025} & $10.044 \pm 0.084$ & $0.375 \pm 0.000$ & $0.348 \pm 0.012$ & $141.341 \pm 11.406$ & $\mathbf{0.582 \pm 0.015}$ & $31.568 \pm 1.668$ \\
U-Shape Transformer~\cite{peng2023u} & $9.282 \pm 0.097$ & $0.279 \pm 0.013$ & $0.380 \pm 0.007$ & $166.642 \pm 6.652$ & $0.794 \pm 0.011$ & $35.520 \pm 1.463$ \\
KAUSTics-Net (ours) & $9.652 \pm 0.139$ & $0.291 \pm 0.022$ & $0.374 \pm 0.006$ & $165.348 \pm 8.477$ & $0.671 \pm 0.067$ & $33.944 \pm 0.388$ \\
\hline
\multicolumn{7}{l}{\textbf{No sunglint}} \\
\hline
Unprocessed input & $10.492$ & $0.374$ & $0.342$ & $142.779$ & $0.643$ & $32.785$ \\
CycleGAN~\cite{zhu2017unpaired} & $10.185 \pm 0.142$ & $\mathbf{0.360 \pm 0.013}$ & $\mathbf{0.316 \pm 0.005}$ & $\mathbf{104.040 \pm 3.521}$ & $0.690 \pm 0.031$ & $29.061 \pm 0.081$ \\
UGAN~\cite{fabbri2018enhancing} & $\mathbf{10.310 \pm 0.125}$ & $0.334 \pm 0.006$ & $0.322 \pm 0.004$ & $114.332 \pm 2.521$ & $0.758 \pm 0.016$ & $\mathbf{28.110 \pm 0.565}$ \\
FUnIE-GAN~\cite{islam2020fast} & $9.405 \pm 0.153$ & $0.217 \pm 0.012$ & $0.372 \pm 0.018$ & $160.120 \pm 19.905$ & $0.736 \pm 0.013$ & $34.553 \pm 1.841$ \\
DeWaterGAN~\cite{10720144} & $10.278 \pm 0.232$ & $0.354 \pm 0.013$ & $0.343 \pm 0.009$ & $138.300 \pm 7.346$ & $0.699 \pm 0.043$ & $31.775 \pm 1.076$ \\
DiffWater~\cite{guan2023diffwater} & $9.832 \pm 0.329$ & $0.244 \pm 0.037$ & $0.333 \pm 0.011$ & $117.792 \pm 6.896$ & $0.739 \pm 0.056$ & $29.726 \pm 3.001$ \\
ResShift~\cite{yue2023resshift} & $9.804 \pm 0.260$ & $0.291 \pm 0.013$ & $0.354 \pm 0.009$ & $143.111 \pm 8.129$ & $\mathbf{0.643 \pm 0.022}$ & $31.938 \pm 0.703$ \\
SS-UIE~\cite{peng_ssuie_2025} & $10.073 \pm 0.073$ & $0.343 \pm 0.002$ & $0.346 \pm 0.011$ & $139.535 \pm 10.727$ & $0.647 \pm 0.013$ & $31.339 \pm 1.582$ \\
U-Shape Transformer~\cite{peng2023u} & $9.220 \pm 0.084$ & $0.269 \pm 0.011$ & $0.382 \pm 0.006$ & $168.495 \pm 6.251$ & $0.830 \pm 0.011$ & $35.760 \pm 1.436$ \\
KAUSTics-Net (ours) & $9.632 \pm 0.147$ & $0.268 \pm 0.019$ & $0.374 \pm 0.004$ & $165.016 \pm 7.423$ & $0.729 \pm 0.062$ & $33.948 \pm 0.270$ \\
\hline
\end{tabular}%
}
\end{table*}

\subsection{No-Reference Image Quality Results}
Table~\ref{tab:supp_zeroshot_blind} reports nine no-reference metrics for the same restored outputs and checkpoints used in the full-reference evaluation. The metrics assess statistical or learned perceptual image properties without access to the reference image and cannot establish recovery of obscured seabed content. Sharpness and entropy describe changes in spatial intensity structure and are not indicators of mapping accuracy.

\begin{table*}[!t]
\centering
\caption{Sea-Undistort zero-shot no-reference metrics on the same 120 held-out scenes per configuration and the same checkpoints as Table~\ref{tab:supp_zeroshot_full}. Results are mean $\pm$ sample standard deviation over three source-training seeds. Sharpness and entropy are descriptive diagnostics and are not ranked.}
\label{tab:supp_zeroshot_blind}
\footnotesize
\setlength{\tabcolsep}{3pt}
\renewcommand{\arraystretch}{1.12}
\resizebox{\textwidth}{!}{%
\begin{tabular}{lccccccccc}
\hline
\textbf{Method} & {\bfseries\boldmath NIQE $\downarrow$} & {\bfseries\boldmath BRISQUE $\downarrow$} & {\bfseries\boldmath PIQE $\downarrow$} & {\bfseries\boldmath ILNIQE $\downarrow$} & {\bfseries\boldmath LIQE $\uparrow$} & {\bfseries\boldmath CLIPIQ $\uparrow$} & {\bfseries\boldmath MUSIQ\_AVA $\uparrow$} & {\bfseries\boldmath Sharpness} & {\bfseries\boldmath Entropy} \\
\hline
\multicolumn{10}{l}{\textbf{All effects}} \\
\hline
Unprocessed input & $8.348$ & $30.378$ & $23.864$ & $64.283$ & $1.935$ & $0.388$ & $3.205$ & $1380.087$ & $5.211$ \\
CycleGAN~\cite{zhu2017unpaired} & $\mathbf{6.710 \pm 0.286}$ & $20.166 \pm 6.844$ & $\mathbf{10.621 \pm 1.945}$ & $74.590 \pm 6.220$ & $1.525 \pm 0.120$ & $0.325 \pm 0.043$ & $3.072 \pm 0.033$ & $191.879 \pm 47.371$ & $5.257 \pm 0.294$ \\
UGAN~\cite{fabbri2018enhancing} & $7.389 \pm 0.166$ & $26.358 \pm 2.449$ & $16.154 \pm 0.495$ & $98.233 \pm 0.339$ & $1.033 \pm 0.004$ & $0.193 \pm 0.004$ & $2.688 \pm 0.028$ & $106.640 \pm 14.310$ & $5.379 \pm 0.106$ \\
FUnIE-GAN~\cite{islam2020fast} & $10.580 \pm 0.142$ & $\mathbf{20.007 \pm 12.755}$ & $37.940 \pm 7.249$ & $112.441 \pm 3.868$ & $1.165 \pm 0.066$ & $0.301 \pm 0.036$ & $3.075 \pm 0.065$ & $708.780 \pm 432.149$ & $5.732 \pm 0.297$ \\
DeWaterGAN~\cite{10720144} & $9.392 \pm 0.626$ & $42.390 \pm 3.089$ & $39.822 \pm 4.841$ & $79.853 \pm 3.556$ & $1.629 \pm 0.071$ & $0.296 \pm 0.026$ & $2.904 \pm 0.019$ & $428.506 \pm 51.456$ & $4.782 \pm 0.126$ \\
DiffWater~\cite{guan2023diffwater} & $6.746 \pm 0.843$ & $24.290 \pm 3.720$ & $26.867 \pm 7.679$ & $108.667 \pm 3.907$ & $1.132 \pm 0.081$ & $0.352 \pm 0.028$ & $2.900 \pm 0.113$ & $902.127 \pm 409.353$ & $5.845 \pm 0.310$ \\
ResShift~\cite{yue2023resshift} & $11.847 \pm 0.147$ & $36.287 \pm 1.034$ & $41.171 \pm 3.754$ & $77.343 \pm 7.123$ & $\mathbf{2.012 \pm 0.094}$ & $\mathbf{0.443 \pm 0.022}$ & $3.057 \pm 0.049$ & $1004.803 \pm 194.482$ & $5.579 \pm 0.137$ \\
SS-UIE~\cite{peng_ssuie_2025} & $9.482 \pm 0.142$ & $33.114 \pm 0.299$ & $27.912 \pm 2.082$ & $\mathbf{67.046 \pm 0.497}$ & $1.618 \pm 0.033$ & $0.350 \pm 0.016$ & $3.106 \pm 0.057$ & $917.371 \pm 66.677$ & $5.341 \pm 0.030$ \\
U-Shape Transformer~\cite{peng2023u} & $10.372 \pm 0.318$ & $55.152 \pm 2.170$ & $56.650 \pm 3.731$ & $110.763 \pm 2.429$ & $1.594 \pm 0.058$ & $0.323 \pm 0.023$ & $2.517 \pm 0.008$ & $54.806 \pm 17.372$ & $4.573 \pm 0.247$ \\
KAUSTics-Net (ours) & $8.851 \pm 1.358$ & $33.022 \pm 11.379$ & $37.992 \pm 7.207$ & $67.477 \pm 1.602$ & $1.649 \pm 0.087$ & $0.389 \pm 0.033$ & $\mathbf{3.119 \pm 0.037}$ & $1170.985 \pm 437.053$ & $5.898 \pm 0.144$ \\
\hline
\multicolumn{10}{l}{\textbf{No waves}} \\
\hline
Unprocessed input & $7.769$ & $19.143$ & $18.021$ & $61.298$ & $1.690$ & $0.390$ & $3.148$ & $346.365$ & $4.682$ \\
CycleGAN~\cite{zhu2017unpaired} & $7.333 \pm 0.377$ & $\mathbf{15.951 \pm 5.577}$ & $\mathbf{10.106 \pm 1.252}$ & $71.704 \pm 5.723$ & $1.430 \pm 0.057$ & $0.327 \pm 0.047$ & $3.139 \pm 0.057$ & $139.709 \pm 52.918$ & $4.982 \pm 0.329$ \\
UGAN~\cite{fabbri2018enhancing} & $7.651 \pm 0.382$ & $20.141 \pm 2.647$ & $16.306 \pm 0.606$ & $92.603 \pm 0.280$ & $1.023 \pm 0.003$ & $0.181 \pm 0.006$ & $2.727 \pm 0.032$ & $109.704 \pm 15.542$ & $5.188 \pm 0.132$ \\
FUnIE-GAN~\cite{islam2020fast} & $11.482 \pm 0.281$ & $21.221 \pm 14.968$ & $40.763 \pm 7.208$ & $114.349 \pm 3.197$ & $1.143 \pm 0.069$ & $0.296 \pm 0.049$ & $3.112 \pm 0.064$ & $726.003 \pm 491.676$ & $5.650 \pm 0.363$ \\
DeWaterGAN~\cite{10720144} & $9.568 \pm 0.752$ & $40.850 \pm 4.561$ & $42.802 \pm 6.551$ & $80.916 \pm 5.325$ & $1.407 \pm 0.102$ & $0.274 \pm 0.027$ & $2.854 \pm 0.013$ & $120.345 \pm 22.251$ & $4.338 \pm 0.114$ \\
DiffWater~\cite{guan2023diffwater} & $\mathbf{6.685 \pm 0.897}$ & $23.683 \pm 4.048$ & $28.281 \pm 8.729$ & $108.235 \pm 4.531$ & $1.134 \pm 0.084$ & $0.353 \pm 0.028$ & $2.887 \pm 0.106$ & $874.145 \pm 447.940$ & $5.748 \pm 0.358$ \\
ResShift~\cite{yue2023resshift} & $12.206 \pm 0.142$ & $32.099 \pm 1.502$ & $43.788 \pm 4.362$ & $72.420 \pm 9.134$ & $\mathbf{1.849 \pm 0.164}$ & $\mathbf{0.440 \pm 0.034}$ & $3.071 \pm 0.062$ & $868.939 \pm 262.146$ & $5.371 \pm 0.182$ \\
SS-UIE~\cite{peng_ssuie_2025} & $7.012 \pm 0.408$ & $19.574 \pm 0.212$ & $21.884 \pm 2.951$ & $\mathbf{63.050 \pm 0.915}$ & $1.492 \pm 0.029$ & $0.356 \pm 0.012$ & $3.137 \pm 0.084$ & $276.931 \pm 11.280$ & $4.992 \pm 0.071$ \\
U-Shape Transformer~\cite{peng2023u} & $11.072 \pm 0.496$ & $53.344 \pm 3.060$ & $59.145 \pm 5.724$ & $106.921 \pm 2.515$ & $1.659 \pm 0.053$ & $0.312 \pm 0.038$ & $2.571 \pm 0.019$ & $51.802 \pm 17.002$ & $4.179 \pm 0.262$ \\
KAUSTics-Net (ours) & $8.840 \pm 1.449$ & $32.682 \pm 13.894$ & $43.217 \pm 7.397$ & $64.279 \pm 2.297$ & $1.499 \pm 0.112$ & $0.365 \pm 0.040$ & $\mathbf{3.162 \pm 0.014}$ & $1200.553 \pm 473.620$ & $5.744 \pm 0.182$ \\
\hline
\multicolumn{10}{l}{\textbf{No sunglint}} \\
\hline
Unprocessed input & $6.641$ & $16.585$ & $15.804$ & $64.752$ & $1.610$ & $0.375$ & $3.082$ & $260.025$ & $4.776$ \\
CycleGAN~\cite{zhu2017unpaired} & $6.751 \pm 0.359$ & $18.680 \pm 6.901$ & $\mathbf{9.797 \pm 1.104}$ & $74.923 \pm 7.569$ & $1.419 \pm 0.113$ & $0.325 \pm 0.054$ & $3.097 \pm 0.049$ & $132.703 \pm 69.036$ & $5.118 \pm 0.354$ \\
UGAN~\cite{fabbri2018enhancing} & $7.446 \pm 0.335$ & $19.821 \pm 2.940$ & $16.325 \pm 0.955$ & $93.273 \pm 0.108$ & $1.031 \pm 0.001$ & $0.177 \pm 0.006$ & $2.732 \pm 0.020$ & $105.509 \pm 15.540$ & $5.223 \pm 0.139$ \\
FUnIE-GAN~\cite{islam2020fast} & $11.930 \pm 0.285$ & $22.426 \pm 14.729$ & $42.400 \pm 6.743$ & $120.552 \pm 4.363$ & $1.148 \pm 0.065$ & $0.304 \pm 0.050$ & $3.122 \pm 0.072$ & $679.338 \pm 462.652$ & $5.594 \pm 0.367$ \\
DeWaterGAN~\cite{10720144} & $9.037 \pm 0.914$ & $38.281 \pm 5.174$ & $32.508 \pm 6.957$ & $87.907 \pm 5.598$ & $1.390 \pm 0.119$ & $0.265 \pm 0.022$ & $2.795 \pm 0.035$ & $88.027 \pm 16.363$ & $4.367 \pm 0.136$ \\
DiffWater~\cite{guan2023diffwater} & $6.701 \pm 0.915$ & $23.289 \pm 3.994$ & $28.361 \pm 9.319$ & $108.785 \pm 4.385$ & $1.143 \pm 0.095$ & $0.358 \pm 0.030$ & $2.903 \pm 0.110$ & $837.907 \pm 445.483$ & $5.716 \pm 0.362$ \\
ResShift~\cite{yue2023resshift} & $12.029 \pm 0.122$ & $30.791 \pm 1.645$ & $42.455 \pm 4.505$ & $76.953 \pm 8.880$ & $\mathbf{1.820 \pm 0.130}$ & $\mathbf{0.438 \pm 0.030}$ & $3.098 \pm 0.061$ & $753.065 \pm 225.447$ & $5.336 \pm 0.181$ \\
SS-UIE~\cite{peng_ssuie_2025} & $\mathbf{6.674 \pm 0.176}$ & $\mathbf{18.601 \pm 0.322}$ & $19.738 \pm 2.315$ & $\mathbf{67.029 \pm 0.649}$ & $1.385 \pm 0.032$ & $0.339 \pm 0.023$ & $3.067 \pm 0.081$ & $212.235 \pm 7.728$ & $5.014 \pm 0.058$ \\
U-Shape Transformer~\cite{peng2023u} & $10.492 \pm 0.419$ & $54.236 \pm 2.377$ & $58.881 \pm 4.989$ & $110.092 \pm 3.023$ & $1.803 \pm 0.047$ & $0.329 \pm 0.038$ & $2.557 \pm 0.010$ & $50.389 \pm 16.266$ & $4.173 \pm 0.269$ \\
KAUSTics-Net (ours) & $9.095 \pm 1.455$ & $32.153 \pm 13.465$ & $42.712 \pm 7.642$ & $67.458 \pm 2.501$ & $1.500 \pm 0.116$ & $0.363 \pm 0.041$ & $\mathbf{3.174 \pm 0.025}$ & $1047.934 \pm 422.985$ & $5.660 \pm 0.186$ \\
\hline
\end{tabular}%
}
\end{table*}

\section{KAUSTics Dataset Inventory and Appearance Examples}
The dataset inventory contains 32 original UAV-image identifiers and 1,280 reference patches, with 40 patches per source image. Ten distorted variants per reference yield 12,800 input--reference pairs. Each reference is counted once in the spatial-pixel inventory. The original reference patches are $1000\times1000$ pixels, and the experiments resize both references and distorted observations to $256\times256$ pixels. The original-file audit identified one degraded training image stored at $1344\times1344$ pixels; the other 12,799 degraded images are $1000\times1000$ pixels. The source audit inspected 28,160 original and resized image-file headers without modifying the original files.

The 1,280 original reference patches contain 1,280,000,000 spatial pixels. After resizing, the references contain 83,886,080 spatial pixels. The training, validation, and test groups account for 1,040, 120, and 120 reference patches, respectively, corresponding to 1,040,000,000, 120,000,000, and 120,000,000 original spatial pixels. The resized totals are 68,157,440, 7,864,320, and 7,864,320 spatial pixels. These figures represent image sampling volume, not unique geographic coverage, habitat area, or independent scene diversity.

\begin{figure*}
\centering
\includegraphics[width=0.92\linewidth]{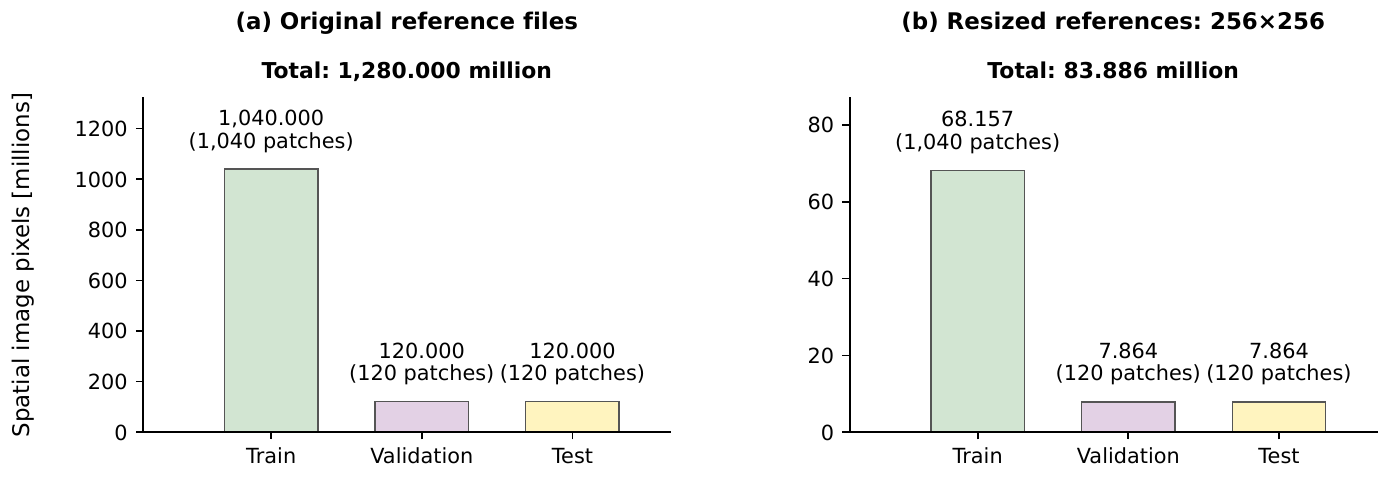}
\caption{Verified reference-image pixel inventory. Training, validation, and test contain 1,040, 120, and 120 reference patches. Original 1000 x 1000 references total 1,280.000 million spatial pixels; resized 256 x 256 references total 83.886 million. References are counted once without their ten degraded observations. Axes have different vertical scales; pixel totals do not measure unique geographic or habitat coverage.}
\end{figure*}

The following image groups describe representative patch appearances from a single shallow-water lagoon: hard-bottom or rock-like texture, sand-dominated appearance, vegetation-like or mixed benthic texture, and dark or low-visibility appearance. These descriptions are visual categories and were not field-validated as substrate classes or species identities. In particular, image darkness and blue tint do not independently establish water depth. Habitat-classification frameworks such as Reef Cover can inform descriptive terminology, but the KAUSTics patches do not include a verified patch-level habitat label manifest. Consequently, category proportions and class-specific pixel counts are not reported.

\begin{figure*}
\centering
\includegraphics[width=0.86\linewidth]{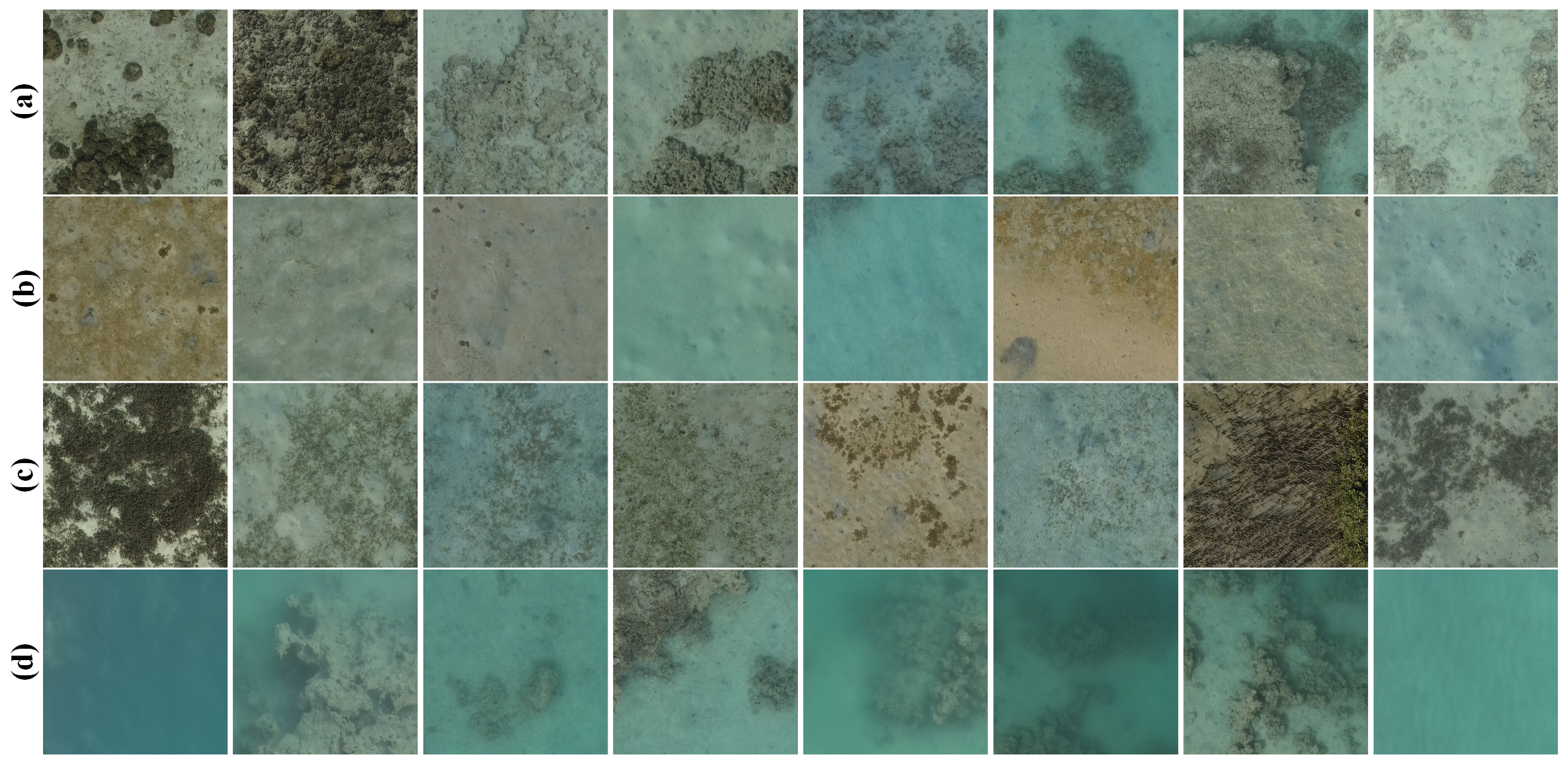}
\caption{Representative calm-water KAUSTics reference patches grouped by descriptive appearance: (a) hard-bottom or rock-like, (b) sand-dominated, (c) vegetation-like or mixed benthic, and (d) dark or low-visibility. These labels are qualitative descriptions and do not indicate field-validated habitat classes, species, or measured water depths.}
\end{figure*}

The corresponding augmented observations illustrate how the same base textures appear under two rendered caustic patterns, two Gaussian-blurred variants, and three empirical color-transfer presets. These transformations introduce different local illumination and radiometric appearances but do not expand the number of independent source scenes. Color presets represent neither measured depth levels nor calibrated light-attenuation coefficients. Visual variation within this survey is not evidence of generalization across independent coastal environments.

\begin{figure*}
\centering
\includegraphics[width=0.86\linewidth]{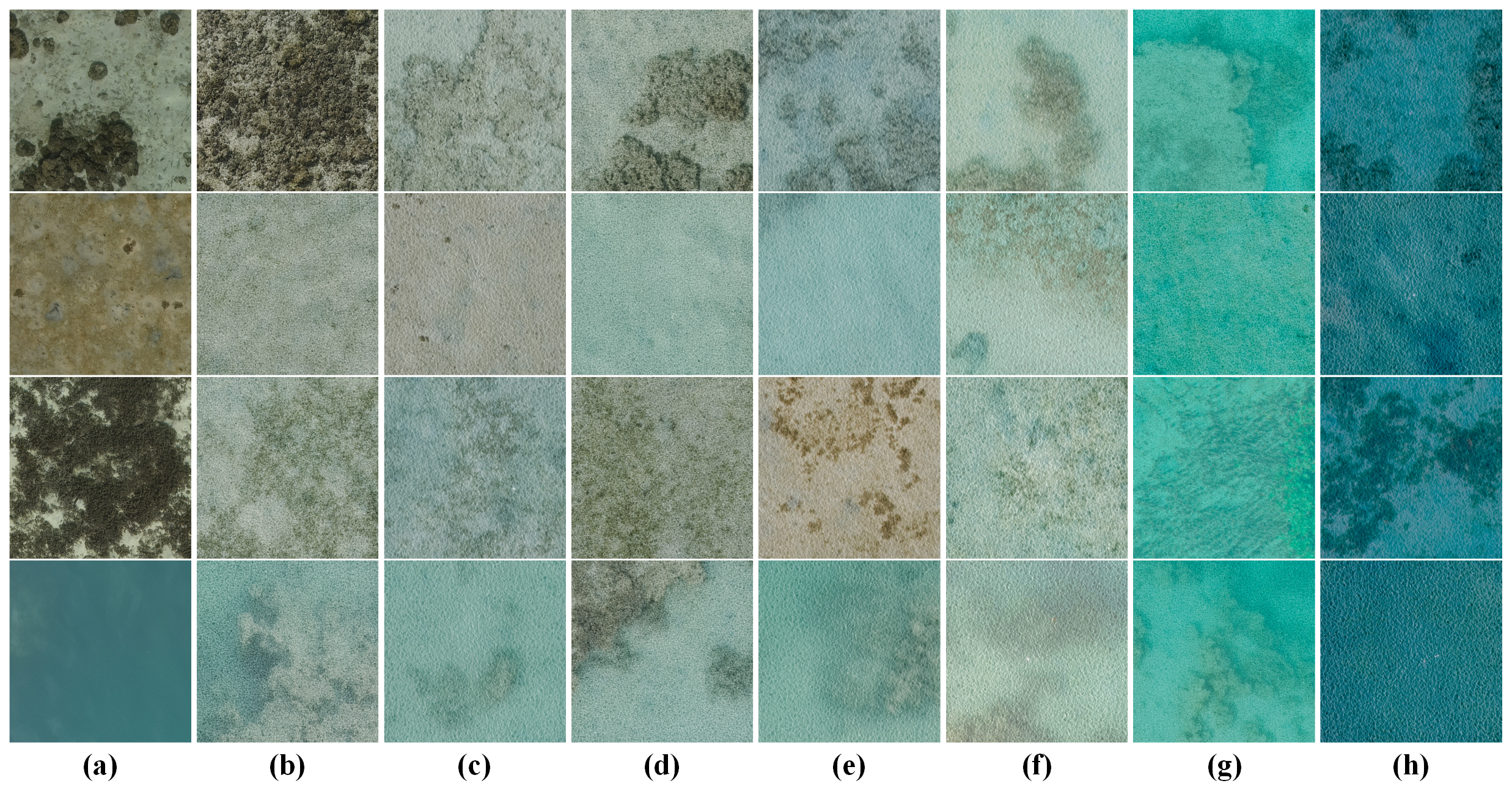}
\caption{KAUSTics references and their rendered or augmented counterparts. Rows follow the descriptive appearance groups of Supplementary Fig. S3. Columns show (a) calm-water references, (b--c) two rendered caustic patterns, (d--e) their Gaussian-blurred variants, and (f--h) empirical color-transfer presets. Color presets are not physical depth classes.}
\end{figure*}
\clearpage
\endgroup
\end{document}